\documentclass[mnsc]{informs4_arxiv}

\usepackage{eqndefns-left} 

\OneAndAHalfSpacedXII

\usepackage{endnotes}
\usepackage{longtable}
\let\footnote=\endnote
\let\enotesize=\normalsize
\def\notesname{Endnotes}%
\def\makeenmark{$^{\theenmark}$}
\def\enoteformat{\rightskip0pt\leftskip0pt\parindent=1.75em
	\leavevmode\llap{\theenmark.\enskip}}

\usepackage{graphicx}
\usepackage{eqndefns-left}
\usepackage{multirow}
\usepackage{hhline}
\usepackage{threeparttable}
\usepackage{pdflscape}
\usepackage[small, margin=1cm]{caption}

\usepackage{appendix}
\usepackage{color}
\definecolor{strcolor}{rgb}{0.6, 0.2, 0.6}
\definecolor{commentcolor}{rgb}{0.3125, 0.5, 0.3125}
\definecolor{keycol}{rgb}{0, 0, 1}

\usepackage{bbm}

\usepackage{listings}
\usepackage{hyperref}
\usepackage{url}

\newcommand {\bea}{\begin{eqnarray}}
	\newcommand {\eea}{\end{eqnarray}}

\def\blot{\quad \mbox{$\vcenter{ \vbox{ \hrule height.4pt
				\hbox{\vrule width.4pt height.9ex \kern.9ex \vrule width.4pt}
				\hrule height.4pt}}$}}

\usepackage[
  backend=bibtex,        
  style=authoryear,      
  citestyle=authoryear,  
  maxcitenames=2,        
  maxbibnames=99,        
  uniquename=false,      
  uniquelist=false,      
  sortcites=true         
]{biblatex}

\renewcommand*{\bibfont}{\small}
\DeclareDelimFormat{multicitedelim}{\addcomma\space}

\DeclareDelimFormat{finalnamedelim}{\addspace\bibstring{and}\space}
\DeclareDelimFormat{multinamedelim}{\addcomma\space}

\TheoremsNumberedThrough     
\ECRepeatTheorems

\EquationsNumberedThrough    

\gdef\AQ#1{}
\gdef\CQ#1{}

\usepackage{graphicx}
\usepackage{float}
\usepackage{outlines}
\usepackage{enumitem}
\usepackage{algorithm}
\usepackage{subcaption}

\usepackage{tablefootnote}
\usepackage{algpseudocode}
\usepackage{xcolor}
\usepackage[most]{tcolorbox}
\usepackage[section]{placeins}
\usepackage{adjustbox}
\usepackage{multirow}
\usepackage{pdflscape}
\usepackage{subcaption}
\usepackage{array}
\usepackage{booktabs}
\usepackage{appendix}
\usepackage{makecell}  

 \usepackage{import}
 
\begin{document}
	

\def\COPYRIGHTHOLDER{INFORMS}%
\def\COPYRIGHTYEAR{2024}%
\def\DOI{\fontsize{7.5}{9.5}\selectfont\sf\bfseries\noindent https://doi.org {Word count =  }}


	\RUNAUTHOR{Kohli et~al.} %

	\RUNTITLE{Sequential choice in ordered bundles}

\TITLE{Sequential choice in ordered bundles}


	\ARTICLEAUTHORS{
 
\AUTHOR{Rajeev Kohli, Kriste Krstovski, Hengyu Kuang, Hengxu Lin}
\AFF{Graduate School of Business, Columbia University}


}
	

\ABSTRACT{
Experience goods such as sporting and artistic events, songs, videos, news stories, podcasts, and television series, are often packaged and consumed in bundles. Many such bundles are ordered in the sense that the individual items are consumed sequentially, one at a time. We examine if an individual's decision to consume the next item in an ordered bundle can be predicted based on his/her consumption pattern for the preceding items. We evaluate several predictive models, including two custom Transformers using decoder-only and encoder-decoder architectures, fine-tuned GPT-3, a custom LSTM model, a reinforcement learning model, two Markov models, and a zero-order model. Using data from Spotify, we find that the custom Transformer with a decoder-only architecture provides the most accurate predictions, both for individual choices and aggregate demand. This model captures a general form of state dependence. Analysis of Transformer attention weights suggests that the consumption of the next item in a bundle is based on approximately equal weighting of all preceding choices. Our results indicate that the Transformer can assist in queuing the next item that an individual is likely to consume from an ordered bundle, predicting the demand for individual items, and personalizing promotions to increase demand.}



\AREAOFREVIEW{Marketing.}

\KEYWORDS{Choice modeling, bundling, personalization, Markov models, neural networks, large language models, Transformers.}

 
\maketitle
\section{Introduction}
Bundling involves selling two or more differentiated products as a single package. This strategy can serve various purposes: enabling price discrimination by monopolists \parencite{stigler1968price, adams1976commodity, mcafee1989multiproduct, schmalensee1984gaussian}, reducing transaction costs \parencite{eppen1991bundling, kenney1983economics}, enhancing economies of scope \parencite{baumol1982applied}, segmenting consumers dynamically \parencite{derdenger2013dynamic}, and deterring market entry \parencite{bernheim1990multimarket, nalebuff2004bundling}. In competitive markets, firms have incentives to offer bundled products \parencite{carbajo1990strategic}. If the number of firms exceeds a certain threshold, bundling can lead to higher market prices, benefit firms, and disadvantage consumers \parencite{zhou2017competitive}. 

Many experience goods, such as sporting and artistic events, songs, videos, news stories, podcasts, and television series, are offered and consumed in bundles. A notable feature of these offerings is that they often consist of \textit{ordered bundles}, meaning the items within a bundle appear in a specific order and are typically consumed in sequence, one at a time.

For example, each of the 30 teams in Major League Baseball (MLB) plays 162 games over a six-month regular season. Fans can choose to purchase individual game tickets or buy season tickets that bundle varying numbers of games (9, 12, 16, 20, 41, or 72) played over time. Similar ordered bundles are available for sports such as basketball, football, and hockey, as well as for live theater, opera, ballet performances, and philharmonic orchestras. For instance, the New York Metropolitan Opera offers approximately 200 performances per season. Individuals can purchase different bundles consisting of six to ten specific performances over the season. 

In the preceding examples, the scheduling of events dictates the order of items in each bundle; for instance, game 3 always follows game 1. Similarly, there is a natural time order to the episodes in a television series, even when entire seasons are released simultaneously for binge watching on platforms like Netflix, Amazon Prime, Disney+, or Hulu. However, some digital platforms package content where the same subset of items can appear in different sequences across bundles. For instance, Spotify curates song playlists, YouTube assembles video reels, and the New York Times organizes stories into newsfeeds. Each playlist, video reel, or newsfeed is a bundle formed by selecting a subset from a large number of items. These providers may incorporate overlapping sets of items in varying sequences across different bundles. Depending on the content, ordered bundles can be consumed quickly (e.g., video reels) or over an extended period (e.g., baseball games).

Motivated by these examples, we  examine whether an individual’s decision to consume the next item in an ordered bundle can be predicted based on his/her consumption pattern for the preceding items in the bundle.  Formally, we define the following \textit{sequential prediction problem}. Consider a bundle of $n$ items. On any given occasion, a consumer can decide sequentially whether to consume each item in the bundle (possibly more than once). Suppose we observe a consumer's choices for the first $i$ items, where $0<i<n$. Can we use this observed sequence, together with information on the features of the items (and/or the consumer), to predict whether the consumer will choose item $(i+1)$? Observe that heterogeneity among individuals is modeled by the  differences in the consumption patterns leading up to  the choice of item $i+1$.
 
The ability to predict whether an individual will consume the next item in a bundle can be useful for sellers. For instance, forecasting the number of season ticket holders likely to attend the next performance can assist a philharmonic orchestra in determining how many individual tickets to offer for that performance \parencite{soman2001transaction}. Similarly, predicting whether a user will listen to the next song on a playlist or the next episode of a podcast can help content providers curate the next item in a listener’s queue. Streaming platforms can also benefit from such predictions. For instance, Amazon Prime Video recently acquired the global rights to stream a subset (bundle) of 66 regular-season NBA games starting in the 2025-26 season \parencite{amazon}. By tracking individual viewing behavior for these games, the platform could estimate the likelihood of an individual watching the next game. This capability enables targeted marketing messages to encourage viewership. Additionally, it allows for predictions of overall game viewership across consumer and geographic segments, informs advertising rates, and help select content programming to accompany the game broadcast.

A natural approach to addressing the sequential prediction problem is to use a Markov model. A first-order Markov model conditions the probability of consuming the $(i+1$)st item on whether an individual consumed the $i$th item. Such a model can capture patterns of persistence; for example, consumers might be more likely to select item $i+1$ if they chose item $i$, and less likely if they did not. Allowing the transition probabilities to vary by item can capture the effect of the popularity of item $i+1$, as well as the idiosyncratic effects of consuming or not consuming item  $i$ on the probability of consuming item  $i+1$. A higher-order Markov model generalizes this approach, allowing predictions for item $i+1$ to depend on each of the preceding $2^k$ consumption sequences, where $0<k\le i$. The main limitation of higher-order Markov models is that there may not be enough data to reliably estimate the transition probabilities from all the states. For this reason, most applications of Markov models reported in the literature are limited to first-order models. 

In this paper, we consider an alternative approach using deep learning models explicitly designed for sequential prediction tasks. These models include Long Short-Term Memory networks (LSTMs), a type of recurrent neural network (RNN), and Transformers, which utilize an attention mechanism to weigh preceding outcomes when predicting the next outcome.

Transformers are similar to Markov models in that they predict the next outcome based on one or more preceding outcomes. However, they employ a different approach by using an attention mechanism, which is recognized as central to the performance of large language models (LLMs) like ChatGPT. While the present problem shares a similar structure with LLMs, predicting the choice of the next item in an ordered bundle is much simpler than predicting the next word in a sequence. Unlike the transition probabilities of Markov models, attention patterns can be complex and generally difficult to interpret. We investigate whether they are more interpretable in the context of predicting consumption outcomes.

We develop custom LSTM and Transformer models, the latter using both a decoder-only  architecture and a bidirectional encoder architecture. We also evaluate GPT 3, a pre-trained LLM based on a decoder-only Transformer architecture. The reason for evaluating GPT is that it is a very large model that predicts well in a wide range of contexts, including predicting the next word in a sequence. Our task is similar but much simpler since we only predict a small number of outcomes. It would be useful, both computationally and for ease of use, if a widely used LLM could be fine-tuned to obtain accurate predictions for the next in a sequence of outcomes. The use of LLMs to capture consumer preferences and predict demand has been assessed in recent research by  \textcite{goli2024frontiers}, \textcite{brand2023using} and \textcite{gui2023challenge}. 

We use data from Spotify, a music-streaming service, to evaluate our models. The service allows listeners to stream musical playlists. Each playlist is an ordered bundle of a small number of songs (items). Our data span multiple playlists and describe the listening behavior of individuals who played each playlist. Given a playlist with $n$ songs, our models predict whether a person will play, replay or skip song $i+1$ based on the pattern of his/her listening behavior for one or more of the preceding $i$ songs, as well as the features of the next song, $i+1$,  where $i<n$.  

Our results show that the decoder-only Transformer obtains the most accurate predictions, followed by LSTM. The first-order Markov model predicted remarkably well for a  simple model that has very few parameters. If prediction were not the only goal, we would consider it to be a very good model of listening behavior. Fine-tuned GPT provides  the worst predictions, with the caveat that its performance was likely affected by a smaller sample size since the model can only use distinct sequences for fine tuning. A notable finding is that the Transformer attention mechanism displays interpretable patterns. This is not a common feature of the attention mechanism, which is generally difficult to interpret in large models. However, the present model predicts only skip, play or replay, and it becomes possible to assess if similar patterns of attention weights are associated with each of the three outcomes. The Transformer predicts that a person will continue listening to the next track on a playlist when the attention weights are approximately uniformly distributed over all preceding outcomes; and it predicts a lower probability of listening when the weight distribution deviates from the uniform. Overall, our results suggest that self-attention holds significant potential for modeling the effect of past outcomes when predicting sequential consumption patterns. They also suggest that a simple Markov model, which has very few parameters, predict as well as, or better than, several more complex models.

\section{Related research}
The present research is related to the consumer behavior literature concerning the bundling of experience goods. This literature suggests that bundling can cater to variety-seeking behavior \parencite{mcalister1982dynamic}, facilitate the consumption of complementary products \parencite{lewbel1985bundling}, and reduce risk or search costs. \textcite{soman2001transaction} show that a product like a theater performance is less likely to be consumed when purchased as part of a bundle, because the payment for a bundle is decoupled from the subsequent consumption of the individual products. They also show  that as the number of items in a bundle increases, the probability of consuming an individual item decreases. Although not in the context of bundling, \textcite{simonson1990effect} shows that uncertainty about future preferences can result in consumers opting for greater variety among the products they purchase for subsequently consumption over time. \textcite{schweidel2016binge} examine the consumption of bundles comprising episodes of TV series and find that the more episodes a user views, the more likely they are to continue watching the next episode.

The present research is also related to work on state-dependent choices and to an emerging literature on session-aware recommendation. Research on state-dependent choices includes random utility models that predict choices using state variables and their interactions with other variables \parencite{guadagni1983logit}. It also encompasses structural choice models that capture the underlying decision processes and the influence of state variables on utility functions or decision rules \parencite{seetharaman2004modeling, dube2010state}. Additionally, Markov models are used to characterize state dependence, such as in brand-switching behavior \parencite{morrison1982modelling}, transitions between latent customer states \parencite{netzer2008hidden}, customer relationship dynamics \parencite{pfeifer2000modeling}, consumer commitment \parencite{ascarza2013joint}, and social-network interactions \parencite{wang2013modeling}. In the operations literature, Markov models have been applied to problems like assortment planning \parencite{blanchet2016markov} and revenue management \parencite{feldman2017revenue}. In this research, we employ a Markov model to condition the probability of an individual selecting the next item in an ordered bundle on whether he/she selected the preceding item. 

Research on session-aware recommendation systems predicts dynamic user choices over a short period of time \parencite{quadrana2018sequence, wang2021survey}. For example, our empirical application involves predicting whether an individual will replay the last song, play the next song, or skip the next song on a musical playlist (which corresponds to an ordered bundle of songs). A related but distinct problem was presented in a challenge organized by the music streaming service Spotify. Contestants were provided with data on the first half of individual listening sessions across many playlists and were asked to predict which tracks would be skipped in the second half of the sessions \parencite{brost2019music}. The contestants employed various machine learning approaches to infer and predict skip patterns, capturing complex nonlinear relationships between outcomes and predictive features without imposing strict parametric assumptions. \textcite{zhu2019sessionbased} proposed the best performing model in the challenge, achieving a hit rate of 65.1\%. Subsequently, \textcite{meggetto2023people} achieved 82\% hit rate using DQN. They noted that the historical data used in the Spotify Challenge had a data leak because certain features could not be available at the time of the prediction. The most important among these features was the duration of a listening session. The hit rate  of Megetto et al.'s model dropped to 66.4\% when it did not use this variable. \textcite{yin2024meta} used a logit model to predict skip/play behavior using the data from the Spotify Challenge. The parameters of this logit model were learnt using a Transformer in which contextual variables (e.g., session length, song position) and musical features (e.g., beat, acoustic) were employed as inputs. The model achieved 73.2\%  hit rate for the first predicted track, and 62.9\% average hit rate over multiple tracks in holdout sessions. 

As noted, two deep learning models ---  RNNs and Transformers --- are central to the present research. We review these briefly below.  

\medskip\noindent
{\bf Recurrent neural networks (RNNs).}
RNNs are used for sequential prediction tasks. Unlike feedforward neural networks, which process inputs independently, RNNs use the output from the previous time step as part of the input for the current time step. This feature allows RNNs to maintain a form of memory when predicting the next outcome. RNNs  can process arbitrary long sequences of inputs and share the same layer parameters across all inputs.

The most widely used RNN is the long short-term memory (LSTM) network. While the original (Elmar) RNN can access the complete information sequence, element by element, it has limited ability for capturing long dependencies (such as between the first and tenth song in a session). The primary reason for this limitation  is that the model uses the same hidden layer both to capture  recent, local information (short memory) and to propagate distant, past information (long memory). A second reason is that vanishing or exploding gradients can prevent error propagation through the network.  The LSTM addresses these limitations by introducing gated units, which control the amount of information ``forgotten'' from the hidden layer and ``added'' to the memory from the current input \parencite{hochreiter1997long}.

\medskip\noindent
{\bf Transformers.}
\textcite{vaswani2017attention} introduced the Transformer, a deep learning architecture that replaces the recurrent structure in a RNN by an attention mechanism. Transformers were originally used in LLMs, which have very large numbers of parameters and are trained using vast sets of textual data. LLMs can be used to model not only language but also many different modalities, such as image \parencite{dosovitskiy2020image, Yu2023ScalingAM}, video \parencite{arnab2021vivit, alayrac2022flamingo}, and audio \parencite{gong21b_interspeech, chen2024vall-e}.   Notable examples of LLMs are BERT \parencite{devlin2018bert}, GPT \parencite{brown2020language, chatgpt, openai2023gpt4}, Llama  \parencite{touvron2023llamaa,touvron2023llama2,meta2024introducing} and  Gemini \parencite{team2023gemini}. 

Self-attention is a pivotal component of Transformers that has enabled LLMs to achieve exceptional performance. \textcite{vaswani2017attention} describes self-attention as ``$\dots$ an attention mechanism relating different positions of a single sequence in order to compute a representation of the sequence.'' Unlike RNNs, a Transformer concurrently processes all the data in a sequence, thus retaining long-term dependencies and speeding up the computation. The attention mechanism has been further developed to allow attention to data spanning multiple levels of a hierarchy \parencite{yang2016hierarchical} and co-attention to multiple sources \parencite{lu2016hierarchical}. Self-attention weights can be computed using different methods, including an additive model that uses a single-layer feedforward network \parencite{bahdanau2014neural}, dot-product alignment \parencite{luong2015effective}, scaled dot-product \parencite{vaswani2017attention}, and similarity-based attention \parencite{ma2017interactive}. Additionally, spatial attention has been developed for image processing \parencite{dosovitskiy2020image}, temporal attention for sequence data \parencite{li2019global}, and channel attention to focus on specific data channels \parencite{yang2020gated}. These innovations continue to expand the versatility and applicability of Transformers.

The Generative Pre-trained Transformer (GPT) is one of the best-known LLMs, trained on a massive amount of data spanning a wide range of linguistic constructs, topics, and writing styles. GPT excels in text-agnostic tasks such as machine translation, sentence/paragraph completion tasks (i.e., cloze tasks), question answering, text summarization, and Winograd Schema-like tasks \parencite{levesque2012winograd}. Successive generations of GPT have had more parameters, utilized more training data, and demonstrated improving and exceptional performance on natural language processing tasks \parencite{radford2018improving, radford2019language, brown2020language, openai2023gpt4}. They also exhibit emergent capabilities on tasks for which they were not specifically trained \parencite{wei2022emergent}. Recent research has begun exploring the use of LLMs, particularly GPT, in marketing research. For instance, \textcite{li2024frontiers} demonstrate that perceptual analysis using LLMs closely aligns with analyses from human surveys. Studies by \textcite{goli2024frontiers} and \textcite{brand2023using} investigate whether GPT-3.5 \parencite{chatgpt} can effectively understand consumer preferences, and \textcite{gui2023challenge} assess the suitability of GPT-4 \parencite{openai2023gpt4} for demand estimation.

\section{Problem description}
We begin by formalizing the sequential-prediction problem. 

Let $n$ denote the number of items in an ordered bundle. Let $x_i$ denote the number of times an individual chooses the $i$th item. For example, in the application reported later in the paper, $x_i$ represents the number of times an individual listens to song $i$ on a playlist. 

We consider discrete time steps. We say that an individual is in {\it position} $j$ at time step $j$, where $j\ge 1$. An individual enters a position in a {\it state}, makes a choice, enters a new state, and then proceeds to the next position, stopping when no further choice can be made. The choice can be to consume another unit of the last item, to consume a unit of the next item, or to neither consume the last item or the next item in a sequence. Below is a formal description of the process.  

\medskip\noindent
1. An individual is initially in position $j=1$ and decides whether to choose item $i=1$. After making the decision, the individual enters the {\it state} $s_1=(x_1)$, where $x_1=1 (0)$ if she/he chooses (does not choose) item $i=1$. 

\medskip\noindent
2. An individual enters position $j+1$ in state $s_{j}=(x_1, \dots, x_i)$, where $1\le i\le n-1$. The value of $x_k$, where $1\le k\le i$, is equal to the number of units of item $k$ that he/she has previously chosen. If $x_i=0$, the individual decides whether to choose item $i+1$. If $x_i\ge 1$, the individual has the additional option of choosing one more unit of item $i$. After making the decision, the individual enters a state $s_{j+1}$ as follows. 
\begin{itemize}
   \item[(i)] If $x_i=0$, then $s_{j+1}=(x_1, \dots, x_i, x_{i+1})$, where $x_{i+1}=1 (0)$ if the individual chooses (does not choose) item $i+1$.
 \item[(ii)] If $x_i\ge 1$, then  $s_{j+1}=(x_1, \dots, x_i+1)$ if the individual chooses one more unit of item $i$; otherwise, $s_{j+1}=(x_1, \dots, x_i, x_{i+1})$, where $x_{i+1}=1 (0)$ if the individual chooses (does not choose) one unit of item $i+1$.
\end{itemize}

\medskip\noindent
The preceding process terminates when there are no more items left (that is, when the individual has chosen as many units of item $i=n$ as he/she wants). The sequential ordering problem is to predict state $s_{j+1}$, given state $s_j$, for all $1\le j\le n$. 

\medskip
Consider an individual who can choose at most $m$ units of item $i$, where $m\ge 1$. Then there can be at most $mn$ possible positions and $(m+1)+(m+1)^2+\dots + (m+1)^n={m+1\over m}((m+1)^n-1)$ possible states, a number that increases exponentially with the value of $n$. In the empirical application reported later, $m=2$, $n$ ranges between 10 and 20, and the sequential-prediction problem can have billions of possible states. As a result, predicting the transition from one state to the next is a non-trivial problem. 

In the following section, we consider multiple models for solving the problem. The simplest of these greatly reduce the complexity of the problem by assuming zero-order or first-order choice processes. The more complex methods use deep learning in which the position of an item in a bundle, as well as item and individual characteristics, are used to make predictions. These models capture  heterogeneity primarily through differences in the preceding sequence of choices made by an individual. The number of distinct consumption sequences preceding item $i$ is ${m\over m-1}(m^i-1)$, a number that increases with $i$. As a result, heterogeneity in preceding choice sequences increases for later items in an ordered bundle.

\section{Models}
We evaluate three Transformer models (GPT-3, a masked decoder, and BERT) for solving the sequential-prediction problem and compare their predictive performance to six other models: two deep learning models (MLP and LSTM), DQN, two Markov models that assume position-independent  (MC) and position-dependent (pMC) transition probabilities, and a zero-order model. Below, we briefly discuss each model and provide implementation details. To provide context, we describe aspects of the models in terms of the problem considered later in the paper, which concerns predicting if an individual listening to the songs on a playlist will replay a song, play the next song, or skip the next song. 

\subsection{Transformer (decoder) model}
The Transformer proposed by \textcite{vaswani2017attention} consists of both encoder and decoder modules. We use only the decoder module, which processes an input sequence of tokens to generate the next sequence of tokens. In contrast, an encoder-only module generates vector encodings for each element of the input sequence of tokens.

In a typical decoder-only setting, each output token (prediction) is fed back as an input to recursively generate a sequence of tokens. For example, GPT-3, which uses a decoder architecture, generates the next word in a sequence based on the preceding words, then uses the newly generated word to predict the next one. In contrast, we use the actual outcomes of a sequence of positions to predict only one (the next) outcome. After each prediction, the input sequence is updated to include the latest actual outcome rather than the latest predicted outcome. This  method is known as teacher forcing \parencite{williams1989learning}.

\begin{figure}[ht]
  \centering
  \includegraphics[width=1\linewidth]{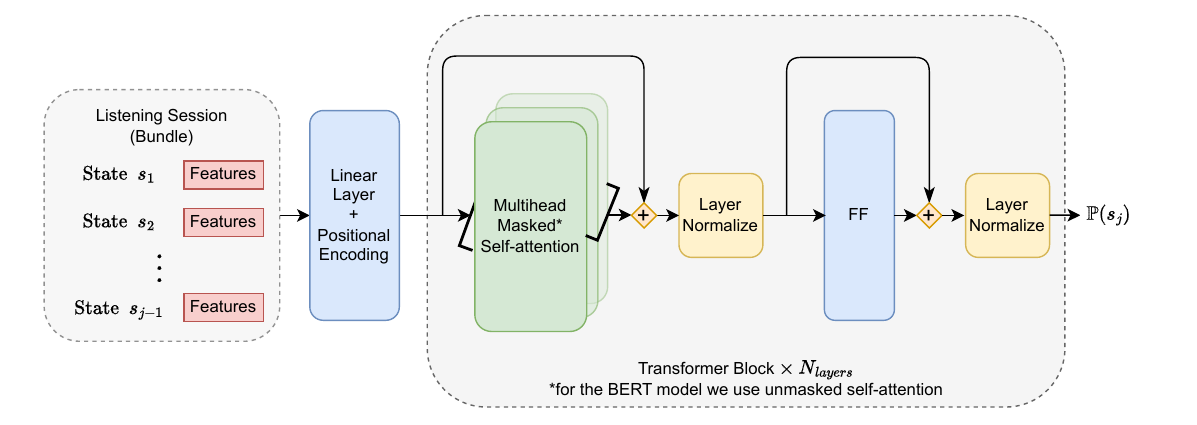}
  \caption{Structure of Transformer used for predicting if a person will replay a song, play the next song, or skip the next song on a playlist.}
  \label{fig:model_tsfm}
\end{figure}

Figure \ref{fig:model_tsfm} shows the steps the Transformer uses to predict the outcome for position $j$ in a session.
As noted, we do not predict the outcome for position $j=1$. The input to the Transformer is a feature vector associated with each position, up and including position $j$. We pass the feature vectors through a linear layer. The output of the linear layer is a set of $j$ embedding vectors $z_1, \dots, z_j$, each of length $d_z=256$. We add fixed positional embeddings (encodings). to the embedding vectors. Each positional embedding is a vector of  length 256. The combined embedding vectors are passed through three Transformer blocks stacked one on top of the other.\footnote{We tested 3, 5, and 6 Transformer blocks and embedding vectors with 128, 256, and 512 elements. Testing used 90\% of randomly selected {\it training data} to estimate model parameters and the remaining 10\% for model validation. The best predictions were obtained using three Transformer blocks and embedding vectors of size 256.
} Each block consists of eight masked self-attention heads; and each head has its own set of query, key and value vectors, each of length $32$. A causal mask is added to ensure that the current position can only attend to itself and previous positions, but not to later positions, when calculating the self-attention scores. The outputs of the eight attention heads are concatenated into a single 256-dimensional attention vector, which is then processed by the remaining layers of the Transformer block comprising layer normalization and a feedforward (FF) layer. Following 
\textcite{vaswani2017attention}, the FF layer consists of two linear transformations.  Each position-wise feedforward network in the FF layer has its own 2048-dimensional hidden layer. Both, the self-attention and feedforward layers have residual connections that add the model input to the model output. Residual connections are standard Transformer components and mitigate the problem of vanishing gradients that can be encountered when training deep learning models.  The output of the final Transformer block is passed through a softmax function that predicts the probability of an outcome for position $j$. We use the PyTorch implementation of the Transformer model \parencite{paszke2019pytorch, pytorch_trans_imp}.

\subsection{Bidirectional Encoder Representations from Transformers (BERT)}
Unlike the decoder, BERT  uses only the encoder module of the Transformer. An encoder generates an output for each input. We ignore all but the last encoder output (that is, the output corresponding to the position for which we wish to make a prediction).  The key differences between BERT and the decoder are: (1) BERT uses learnable instead of fixed positional embeddings --- that is, the model learns a dense, high-dimensional vector that encodes each position (we set the embedding vector length to 256); and (2) BERT uses bidirectional attention --- that is, it does not use the causal mask when calculating self-attention weights. As a result, during the training phase, the self-attention weight for position $j$ is calculated by considering all positions, including the subsequent positions $j+1$ to $n$. However, future information is not available in the prediction phase, and thus is excluded when making predictions.  We employ the Hugging Face version of BERT \parencite{hugging_face_bert} with all default parameters.

\subsection{Generative Pre-trained Transformer (GPT)}
The GPT family of models can be used in zero and few-shot settings for tasks such as text generation. However, our objective is not to generate text but to predict listening behavior, a non-standard task that requires fine-tuning the model using data from listening sessions. At the time of the study, GPT-3 was the most advanced GPT model available. We used its DaVinci variant for fine-tuning. To our knowledge, this paper is the first to report the use of LLMs for predicting music listening behavior. GPT-3 had a limitation that only unique input sequences (listening sessions) could be used for fine-tuning. As discussed later, this constraint significantly reduced the number of listening sessions available for fine-tuning GPT-3.

\begin{figure}[ht]
  \centering
  \includegraphics[width=0.75\linewidth]{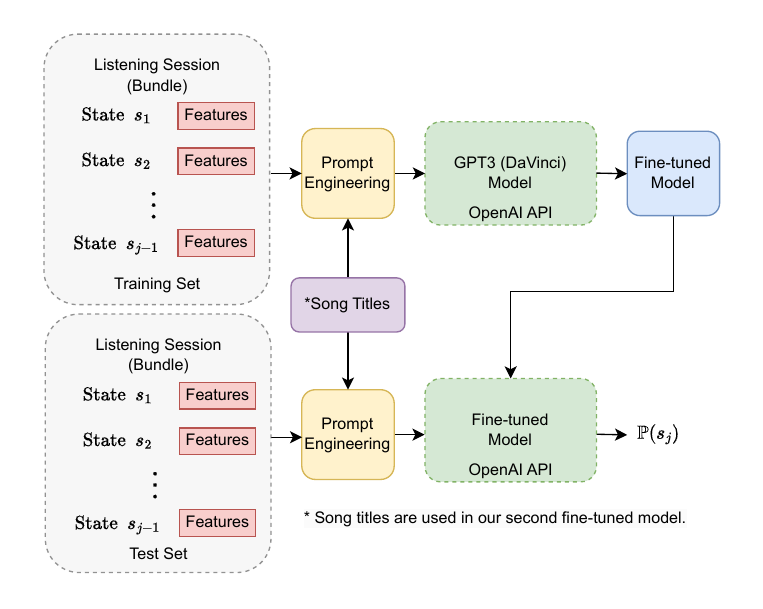}
  \caption{Fine-tuning GPT-3 model (DaVinci) to predict if a person will replay a song, play the next song or skip the next song on a playlist.}
  \label{fig:model_gpt3}
\end{figure}
Figure \ref{fig:model_gpt3} shows the setup for fine-tuning and testing GPT-3. The first step is to represent each song on a playlist by its features. The next step is prompt engineering, followed by fine-tuning, after which the model is used to make predictions for the holdout listening sessions.

\begin{figure}[!htbp]
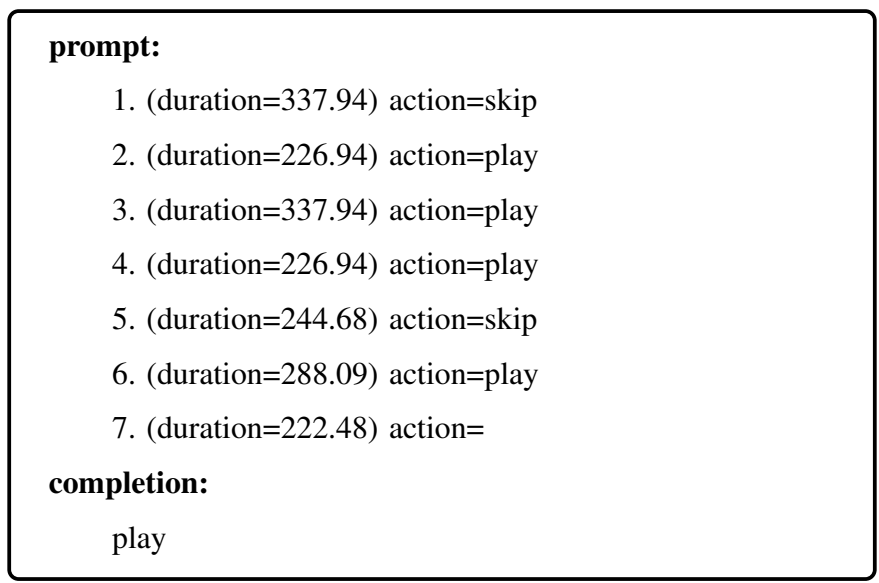

  \centering
\begin{tcolorbox}[fonttitle=\bfseries, colframe=black, colback=white, coltitle=white,halign=left, width=.7\textwidth]
  \textbf{prompt:}\\
 \qquad 1. (duration=337.94) action=skip \\
 \qquad 2. (duration=226.94) action=play \\
 \qquad 3. (duration=337.94) action=play \\
 \qquad  4. (duration=226.94) action=play \\
 \qquad 5. (duration=244.68) action=skip \\
 \qquad 6. (duration=288.09) action=play \\
 \qquad 7. (duration=222.48) action= \\
  \textbf{completion:} \\
  \qquad play\\
\end{tcolorbox}
\caption{An example of a prompt-completion pair used for fine-tuning GPT-3 to predict if a person will replay a song, play the next song or skip the next song on a playlist.}
 \label{fig:prompt_completion_example}
\end{figure}

GPT-3 interacts through prompts and responds with completions. Fine-tuning is performed by providing the model pairs of prompts and completions. A prompt provides information about song features and outcomes (play, replay, or skip) for each of the first $j-1$ positions in a session. The completion is a word (``replay,'' ``play,'' or ``skip'') that specifies the outcome for position $j$.\footnote{\textcite{raffel2020exploring} previously used this method for classification in the context of a Transformer using an encoder-decoder architecture.} 
Figure \ref{fig:prompt_completion_example}  shows an example of a prompt-completion pair for a session with seven positions. The prompt describes the song duration (``duration='') and the user action (``action='') for each of the first six positions. It also provides the duration of the next song but  does not specify the user action. The completion provides the action (i.e. replay the previous song, play the next song, or skip the next song). 

LLM predictions can depend on the natural language descriptions of the task (prompt instructions) and the examples demonstrating the task. Ideally, we would have liked the prompts to include song information (such as song title and artist) to leverage GPT-3’s pre-training on a vast amount of online information. However, such information was unavailable in our dataset.

We experimented with various ways of providing instructions and demonstrations. Each prompt version used different words and phrases but conveyed the same information about song duration and outcome. For example, we experimented with instructions such as, ``Given the following sequence of songs, suggest the next song:'', ``My goal is to predict the play/skip/replay pattern in the following Spotify playlist containing a sequence of 7 songs:'', etc.  Different versions of these instructions performed about equally well. As a result, we settled on the shortest of these prompts because the cost of using GPT-3 depends on the total number of tokens used in the prompts.

\subsection{Long Short-Term Memory (LSTM) }
LSTM is a well-known model in the family of RNNs \parencite{hochreiter1997long}.  Given a listening session, LSTM processes each element of the (input) feature vector $z_{j}$ at position $j$ into a hidden state vector $h_j$ and a cell state vector $c_j$ by recursively leveraging the current feature and the previous hidden state:
$$o_j, h_{j}, c_j =\text{LSTM}(z_{j}, h_{j-1}, c_{j-1}),$$
where $o_j$ is the model output. The action probability is calculated by feeding $o_{j}$ into a simple feedforward module followed by a softmax function. Following \textcite{meggetto2023people}, our LSTM configuration uses two hidden layers, each with 128 units. We used PyTorch's implementation of the LSTM model \parencite{pytorch_lstm_imp}.  

\subsection{Multilayer Perceptron (MLP)}
We employ a three-layer fully connected feedforward neural network, commonly known as an MLP, to serve as a cross-sectional benchmark.  The output of the model is the probability of an outcome for a given position in a session, generated using a softmax function. To maintain comparable model complexity, we configure the MLP with three hidden layers, each containing 256 units. We used PyTorch to implement the MLP \parencite{goodfellow2016deep}.

\subsection{Deep Q-Network (DQN)}
\textcite{meggetto2023people} use DQN, an RL model, to predict skip or play for the Spotify Challenge. Each state is associated with the time a person makes a skip or play decision. States are described using song features. The reward is one if the action matches the actual outcome, and zero otherwise. Predicting an incorrect action terminates an episode during training. The model uses Q-learning \parencite{watkins1989learning}, a model-free approach to RL. The value function is approximated by a three-layer MLP with 128 hidden units.

Conceptually, RL is not appropriate for sequential prediction because its objective is to identify a policy that maximizes a reward function. In our problem, the objective is not to find an optimal policy, and there is no exploration of the state space to find an optimal policy. However, we use DQN as a benchmark model because it achieved the best performance in the Spotify Challenge \parencite{meggetto2023people}. We modify the source code provided by \textcite{dqn_imp} to allow replays and remove the restriction that an episode terminates when the model makes an incorrect prediction in a session. Instead, we change the reward function so that a correct prediction earns a reward of +1 and a false prediction earns a reward of -1. This change utilizes all the data in a listening session for training and substantially improves the model’s hit rate. We estimate the value function using a Q-Network with 256 hidden units per layer, which is comparable to the layer sizes for LSTM, Transformer, and BERT.

\subsection{Markov models}
We consider two first-order Markov models. The first model (MC) assumes the same transition probabilities for each position on a playlist. The second model (pMC) allows position-specific transition probabilities. These transition probabilities are constrained to ensure that (1) the probability of a replay following a skip is zero (a listener cannot skip and then replay a song), and (2) the process terminates after the last song (skipping or playing the next song is infeasible after the last song on a playlist). Our data includes at most one replay per song, and there are playlists where all replays occur for the last song (and thus no transitions from replaying an earlier song to skipping or playing the next song).

\subsection{Zero-order model}
A zero-order model cannot directly model replays, which are inherently conditional outcomes. However, we can infer replays from a zero-order model as follows. Let $p(x_i)$ the probability that song $i$ is played $x_i$ times, where $x_i\ge 0$ is an integer. Then $p(x_i=0)$ is the probability of skipping song $i$, $p(x_i\ge 1)=1-p(x_i=0)$ is the probability of playing song $i$, and $p(x_i\ge 2)$ is the probability of replaying song $i$ at least once. In our dataset, no song is played more than twice and thus $p(x_i\ge 2)=p(x_i=2)$. This simple model can obtain a high hit rate when the probability of an outcome (skip, play, or replay) is high. For example, Table 1 shows that 77\% of all songs on playlist 3 are skipped. Predicting that all listeners skip all songs on the playlist would thus achieve a 77\% hit rate. Similarly, predicting that all listeners play all songs on playlist 9 would achieve a 64\% hit rate. A zero-order model can achieve a higher hit rate by making these predictions separately for each song on a playlist. 

\section{Data}
We used data provided by Spotify, a music streaming service. The dataset contains information on the playlist listening behavior of individuals and was made available by the company for a Sequential Skip Prediction Challenge \parencite{brost2019music}. Each playlist represents an ordered bundle of a small number of (say, $n$) songs. An individual listening to a playlist can skip a song or play it one or more times. Suppose we have observed an individual's listening behavior for the first $i$ songs on a playlist, where $0\le i\le n-1$. Our objective is to predict if he/she will replay song $i$, play song $(i+1)$, or skip song $i+1$, based on the individual's listening pattern for the first $i$ songs  as well as information on the features of the  $(i+1)$st song (we do not have information on the characteristics of individuals, which, if available, could also have been used for making the predictions).

The full dataset contains the logs of more than 130 million listening sessions, metadata for each track, and data on user interactions with the tracks. A session log is a sequence of tracks; track metadata includes song features and popularity; user-interaction data specify if a track was skipped, paused, or played, or if a listener switched to another playlist. We inferred replays from a variable labeled  ``hist\_user\_behavior\_reason\_start''.

\begin{table}[!htbp] \centering 
\caption{Summary statistics for the ten selected playlists.} 
\label{data_desc} 
\begin{tabular}{@{\extracolsep{5pt}} rrrrrrrr} 
\hline
Playlist &  
\begin{tabular}[c]{@{}c@{}}\# of \\ tracks \end{tabular}  & 
\begin{tabular}[c]{@{}c@{}}\# of \\ sessions \end{tabular} &
\begin{tabular}[c]{@{}c@{}}Avg. listening\\ time [min:sec]\end{tabular} & 
\begin{tabular}[c]{@{}c@{}}Avg. \# of \\ played songs  \end{tabular} & 
\begin{tabular}[c]{@{}c@{}}Skip [\%]\end{tabular} & 
\begin{tabular}[c]{@{}c@{}} Play [\%]\end{tabular} & 
\begin{tabular}[c]{@{}c@{}} Replay [\%]\end{tabular} \\
\hline
$1$ & $9$ & $763$ & $16:01$ &   $3.69$ & $59.00$ & $38.96$ & $2.03$ \\ 
$2$ & $9$ & $629$ & $18:30$ & $3.92$ & $56.47$ & $41.39$ & $2.14$ \\ 
$3$ & $19$ & $595$ & $16:27$ &   $4.34$ & $77.15$ & $22.50$ & $0.34$ \\ 
$4$ & $19$ & $529$ & $19:30$ & $5.34$ & $71.87$ & $27.47$ & $0.66$ \\ 
$5$ & $18$ & $526$ & $21:23$ & $5.56$ & $69.10$ & $30.44$ & $0.47$ \\ 
$6$ & $10$ & $482$ & $15:01$ &   $3.64$ & $63.60$ & $35.74$ & $0.66$ \\ 
$7$ & $8$ & $479$ & $20:05$ & $4.26$ & $46.72$ & $50.21$ & $3.07$ \\ 
$8$ & $9$ & $473$ & $22:34$ & $4.91$ & $45.44$ & $51.64$ & $2.92$ \\ 
$9$ & $13$ & $466$ & $36:59$ & $8.85$ & $31.92$ & $66.27$ & $1.82$ \\ 
$10$ & $18$ & $450$ & $18:05$ & $4.63$ & $74.28$ & $25.40$ & $0.32$ \\ 
\hline
\hline
Avg.& $13.20$ & $539.20$ & $20:11$ & $4.93$  & $62.60$ &  $36.19$ &  $1.21$ \\ 
\hline
\end{tabular} 
\end{table}

We selected ten playlists with the highest number of listening sessions. Table \ref{data_desc} provides summary statistics. The number of tracks per playlist ranges between 10 and 20 (average = 13.20), and the number of sessions per playlist between 450 and 763 (average = 539.2). The average song length is approximately three and a half minutes, and the average listening time for a playlist ranges between about 16 minutes and 37 minutes (average = 20 minutes, 11 seconds). On average, listeners play between 3.69 and 8.85 songs on a playlist (the average number of songs across playlists is 4.93). The percentage of skipped songs ranges between about 32\% to 77\% (average = 62.6\%), and replays between 0.32\% and 3.07\% (average = 1.21\%). Playlist 9 stands apart from the others: it has the longest average listening time (just under 37 minutes),  the highest average number of songs played (8.85) and the lowest percentage of skipped songs (31.92\%).   

\begin{figure}[!htbp]
  \centering
  \includegraphics[width=.75\textwidth]{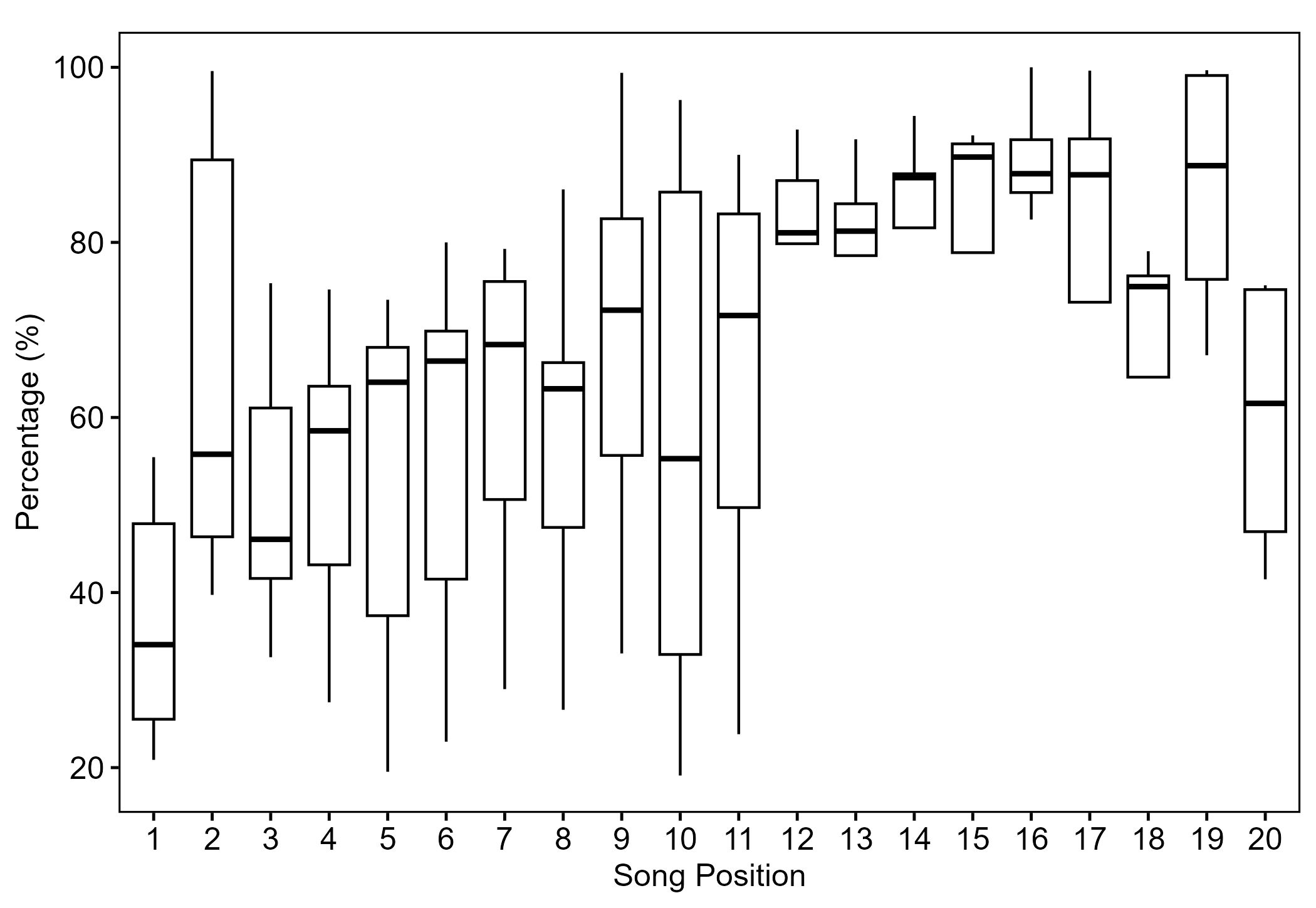}
  \caption{Box plots of skip rates by position across playlists. 
 } \label{fig:data_summary}
\end{figure}

Figure \ref{fig:data_summary} shows the skip rates by position across playlists. The average skip rate is the lowest for the first position, after which it roughly increases, with notable drops at positions 3, 10, 18 and 20. Figure \ref{fig:summary_all} (see Appendix A) plots the skip rates by position for individual playlists. It shows that each playlist has at least one position with nearly 100\% skip rate; that playlists 7, 8, and 9 have low skip rates for most positions; and that playlists 3 and 4 have high skip rates for most positions.

Our training sample consists of 90\% randomly selected listening sessions for each playlist, with the remaining 10\% reserved for model testing. Summary statistics for both the training and test samples are presented in Appendix B. Table \ref{training_sample_size} shows the number of sessions per playlist that were used for fine-tuning GPT-3 and training all other models. The training sample for GPT-3 is substantially smaller because it only allows using unique sessions for fine-tuning. This factor can significantly hamper the performance of GPT-3 compared to the other models. 

    \begin{table}[ht]
      \caption{Number of training sessions per playlist.}
      \label{training_sample_size}
      \centering
      \begin{threeparttable}
      \begin{tabular}{lrrrrrrrrrr}
        \toprule
        & \multicolumn{10}{c}{Playlist}       \\ \cline{2-11} 
        & 1   & 2   & 3   & 4   & 5   & 6   & 7   & 8   & 9   & 10  \\ \cline{2-11} 
      GPT-3& 150 & 184 & 283 & 309 & 295 & 149 & 153 & 165 & 179 & 236 \\
      All other models& 686 & 566 & 535 & 476 & 473 & 433 & 431 & 425 & 419 & 405 \\
      \hline
      \end{tabular}
      \end{threeparttable}
      \end{table}

\section{Analysis and results}
We use the training data to estimate the probability that a song (item) on a playlist will be skipped, played, or replayed during a listening session (ordered bundle). We apply a max-probability rule, predicting the outcome with the highest probability of occurrence for each position. By comparing the predicted and actual outcomes, we calculate the hit rate, which is the percentage of correct predictions across holdout sessions for each song on each playlist. This measure is widely employed to evaluate model performance and was previously used in the Spotify Challenge \parencite{brost2019music}.

Our dataset does not record when a person stops listening to songs on a playlist. Therefore, we need to make an assumption about how many songs to consider in each session when making predictions. One possible assumption is that a person listens to all the songs on a playlist. Another is that a person listens only up to the last song he/she actually heard. To illustrate the difference, imagine a playlist with ten songs, and a session record showing that the fifth song was played but not any subsequent song. Under the first assumption, we consider all ten positions for model training, marking the last five positions as ``skipped.'' Under the second assumption, we truncate the training data after the fifth position. We estimate the models under both assumptions. The second assumption yields slightly lower hit rates for all models, but the overall pattern of results remains unchanged under either assumption. Here, we present the results assuming that each person considers listening to all songs on a playlist. Table \ref{tab:overall_accuracy_truncated} in Appendix C shows the hit rates for each playlist and across playlists assuming data truncation after the last song played in a session. 

Next, we discuss the selection of input features, compare the hit rates across models, examine the accuracy of predictions for different outcomes, contrast the distribution of hit rates across models, and evaluate the effect of including the actual listening time (a data leak) as a model feature.

\subsection{Feature selection}\label{feature_selection}
The first column of Table \ref{tab:feat_importance_full} lists the features available in the dataset for the Spotify Challenge.\footnote{For a description of these variables, see {\footnotesize   \url{https://www.aicrowd.com/challenges/spotify-sequential-skip-prediction-challenge/dataset_files}}} Additionally, it includes a feature called predicted remaining time, which was inferred as follows. Let $\tau_j$ denote the  listening time (i.e., the time songs are played and replayed) starting at position $j$ and ending at the last position in a session. The predicted remaining time for position $j$ is then calculated as the average value of $\tau_j$ across all the sessions associated with a given playlist. We used the training data to estimate the predicted remaining time for each position in a session. To our knowledge, predicted remaining time has not been previously used as a feature for predicting music-listening behavior. 

\begin{table}[htbp]
\centering
\caption{\label{tab:feat_importance_full}Assessing feature importance.}    
    \begin{tabular}{l|rrr}
    \hline
    \multirow{2}{*}{\begin{tabular}[c]{c@{}} Feature \end{tabular} } &  \multicolumn{3}{c}{Importance Metric}  \\
   \cline{2-4}
     &Split & Info. Gain & Shapley Value \\
    \hline
    Previous action  & 68.00 & 5956.43 & 3.04 \\
    Predicted remaining time &  293.00 & 2331.92 & 0.75\\
    US popularity estimate & 160.00 & 7363.85 & -0.09 \\
    Song tempo & 73.00 & 2145.48 & 0.12 \\
    Song length &29.00 & 1215.56 & 0.35 \\
    Song mean dynamic range &  34.00 & 1234.29 & 0.19 \\
    Song key &  15.00 & 188.89 & 0.63  \\
    Short pause before play  & 50.00 & 2563.34 & 0.06 \\
    No pause before play & 61.00 & 683.78 & -0.42\\
    \# of times user seek forward &  10.83 & 89.57 & 1.06\\
    Long pause before play &  9.00 & 27.51 & -0.38 \\
    \# of times user seek backward & 9.46 & 51.30 & 0.45\\
    Hour of day & 206.00 & 722.65 & -1.27 \\
    Song acoustic vector & 12.38 & 464.93 & -0.91 \\
    Song flatness &0.00 & 0.00 & -0.57\\
    Shuffle mode & 0.00 & 0.00 & -0.68 \\
    Song organism &  0.00 & 0.00 & -0.83\\
    Song danceability & 0.00 & 0.00 & -0.86\\
    Song beat strength &  0.00 & 0.00 & -0.91 \\
    Song mechanism &0.00 & 0.00 & -0.92\\
    Song bounciness &  0.00 & 0.00 & -0.92 \\
    Song speechiness & 0.00 & 0.00 & -0.96\\
    Song valence &  0.00 & 0.00 & -0.98\\
    Song liveness &  0.00 & 0.00 & -1.03 \\
    Song acousticness & 0.00 & 0.00 & -1.05 \\
    Song mode & 0.00 & 0.00 & -1.07 \\
    Song energy &  0.00 & 0.00 & -1.11 \\
    Song loudness &0.00 & 0.00 & -1.12 \\
    Song instrumentalness &  0.00 & 0.00 & -1.25 \\
    \hline
    \end{tabular}
\end{table} 

We use the variables shown in Table \ref{tab:feat_importance_full} to train the Transformer and a gradient boosting decision tree (GBDT), which is an ensemble of decision trees \parencite{chen2016xgboost}. We employ the Shapley Value to assess the effect of including a variable on the predictions of the Transformer model \parencite{lundberg2017unified}. We use two measures of feature importance in the GBDT model: (1) the number of splits involving each feature across all decision trees (Split), and (2) the increase in likelihood from all splits of a feature among all trees (Info. Gain). Table \ref{tab:feat_importance_full} shows the values of the three importance metrics. Both previous action and predicted remaining time have a large number of splits, high information gain, and a high Shapley Value.\footnote{``Previous action'' is used 68 times for splitting the decision trees, and the variable increases the likelihood by 5,956.43 after summing the negative likelihood on all splits in the GBDT model. The Shapley Value for ``previous action'' is 3.04, which suggests that it significantly increases the log odds and thus the probability of correct prediction.} Most song features have low scores on the three importance metrics. These results are consistent with earlier findings that previous action is important, but song features are unimportant for predictions in the Spotify Challenge \parencite{meggetto2023people}.

\begin{table}[ht]
\centering 
\begin{threeparttable}
\caption{\label{tab:overall_accuracy_action}Holdout hit rates for alternative sets of predictor variables.}
\begin{tabular}{l|ccccc}
\hline
\multicolumn{1}{l}{\multirow{2}{*}{Feature}}        & \multicolumn{5}{|c}{Avg. Hit Rate}  \\ 
 \cline{2-6}
                                             & Transformer & LSTM & BERT & MLP & DQN                  \\ 
\hline
Predicted remaining time                     & 0.76        & 0.72 & 0.68 &0.71& 0.73                 \\
Previous action                              & 0.83        & 0.80 & 0.79 &0.76 &0.75                 \\
Predicted remaining time and previous action & 0.84        & 0.81 & 0.78 &0.79& 0.78                 \\
All variables in Table \ref{tab:feat_importance_full}                          & 0.85        & 0.79 & 0.76 &0.81& 0.80                 \\
\hline
\end{tabular}
\end{threeparttable}
\end{table}

Next, we trained the Transformer, LSTM, BERT, MLP, and DQN models using (1) all features, and (2) one or both of predicted remaining time and previous action (skip, play, or replay). Table \ref{tab:overall_accuracy_action} presents the hit rate achieved by each model for individual playlists across sessions, and the overall hit rate across all playlists. Limiting the features to only the previous action and predicted remaining time yields  marginal {\it increases} in hit rates for LSTM and BERT models. This finding suggests potential over-parameterization when these models are trained using the full feature set. In contrast, the Transformer model exhibits a slight decrease in hit rate when restricted to these two features. Removing the predicted remaining time feature further results in a minor reduction in hit rates across all five models, including the Transformer. Given these findings, we opted to use only the previous action and predicted remaining time as features.

\subsection{Model predictions}
Table \ref{tab:overall_accuracy} presents the playlist-specific hit rates, and the weighted average hit rate across playlists, for each model (the weights are proportional to the number of observations on the playlists). The table also shows the number of parameters per playlist for each model (the number of parameters varies by playlist for pMC and the zero-order model). We make the following observations based on these results.

\begin{table}[ht]
\caption{Comparison of  hit rates across  models.}
\label{tab:overall_accuracy}
\resizebox{\textwidth}{!}{%
\begin{tabular}{l|r|c|c|c|c|c|c|c|c|c|c|c}
\hline
\multicolumn{1}{l|}{\multirow{2}{*}{Model}} & \multirow{2}{*}{\begin{tabular}[c]{@{}c@{}} \# of \\ Parameters
\end{tabular} } 
& {\multirow{2}{*}{\begin{tabular}[c]{@{}c@{}} Avg. \\ Hit Rate 
\end{tabular} }} 
& \multicolumn{10}{c}{Playlist Hit Rate} \\
   \cline{4-13}
  & & & 1 & 2 & 3 & 4 & 5 & 6 & 7 & 8 & 9 & 10\\
\hline
Transformer & 6.8M & \color{red}{0.84} & 0.80 & \color{red}{0.80} & \color{red}{0.87} & \color{red}{0.87} & \color{red}{0.84} & \color{red}{0.80} & \color{red}{0.77} & \color{red}{0.81} & \color{red}{0.88} & \color{red}{0.84}\\
\hline
LSTM & 1.8M & 0.81 & \color{red}{0.82} & \color{red}{0.80} & 0.82 & 0.85 & 0.82 & 0.77 & 0.72 & 0.78 & 0.82 & 0.79\\
\hline
BERT & 47.6M & 0.78 & 0.73 & 0.71 & 0.82 & 0.83 & 0.78 & 0.79 & 0.71 & 0.74 & 0.73 & 0.80\\
\hline
MLP & 0.2M & 0.79 & 0.74 & 0.74 & 0.83 & 0.84 & 0.80 & 0.78 & 0.70 & 0.75 & 0.79 & 0.79\\
\hline
DQN & 0.2M & 0.78 & 0.75 & 0.74 & 0.83 & 0.84 & 0.79 & 0.78 & 0.70 & 0.69 & 0.79 & 0.79\\
\hline
GPT-3 & Fine-tuned$^{**}$& 0.68 & 0.59 & 0.68 & 0.57 & 0.63 & 0.79 & 0.66 & 0.67 & 0.65 & 0.71 & 0.81\\
\hline
pMC & [45, 95]$^*$  & 0.77 & 0.71 & 0.68 & 0.82 & 0.84 & 0.77 & 0.79 & 0.73 & 0.75 & 0.82 & 0.75\\
\hline
MC & 5  & 0.78 & 0.73 & 0.72 & 0.83 & 0.84 & 0.79 & 0.77 & 0.69 & 0.71 & 0.80 & 0.79\\
\hline
Zero order & [18, 38]$^*$   & 0.70 & 0.66 & 0.61 & 0.76 & 0.76 & 0.71 & 0.69 & 0.61 & 0.63 & 0.58 & 0.73\\
\hline
\end{tabular}%
}
\begin{tablenotes}[para,flushleft]
\footnotesize
\item[] Note: The highest hit rates in each column are shown in red. 
 
 \item[$^{*}$] Ranges for the number of parameters per playlist.
 
 \item[$^{**}$] Pre-trained GPT-3 has 175B parameters. 
 \end{tablenotes}
\end{table}

\begin{enumerate}[label=(\arabic*) ]
    \item  The Transformer correctly predicts the highest percentage (84\%) of outcomes across sessions and playlists. The next best is LSTM (81\%). The Transformer also achieves the highest hit rate for individual playlist, with one exception --- for playlist 1, LSTM obtains a hit rate of 82\%, and the Transformer 80\%.
    BERT uses far more parameters but obtains an average hit rate of only 78\%. Thus, while bi-directional self-attention can produce ``richer'' BERT encodings, this does not translate into better holdout predictions in the present context.
    \item  The two models with the worst average hit rates are GPT-3 (68\%) and the zero-order model (70\%).  Fine-tuned GPT-3 obtains the lowest hit rate across models for playlists 1, 3, 4 and 6, and the zero-order model for playlists 2, 5, 7, 8 and 9. We expected the zero-order model to perform poorly for these playlists because  they have skip rates close to 50\% (its performance improves when a playlist has higher skip or play rates; see Figure \ref{fig:summary_all}).  However, the result for GPT-3 is surprising given its strong capabilities in few-shot learning for tasks such as translation and question-answering \parencite{brown2020language}.\footnote{Additional experiments using a standalone data split of sessions revealed that GPT-3 has almost zero predictive accuracy in zero-shot settings, indicating that fine-tuning does improve its predictions.} The poor present performance of GPT-3 can be attributed to at least two factors. First, predictions are likely to suffer when the pre-trained model has little task-specific knowledge. Our dataset lacks ``real-world'' information, such as song titles and artist names, which could have improved GPT-3's predictions. Second, the restriction that GPT-3 can only use unique data sequences for fine-tuning not only reduces the size of the training sample (see Table \ref{training_sample_size}) but also prevents the model from learning patterns in repeated session. 
    \item As discussed in section 4.6, we extend \textcite{meggetto2023people}'s DQN formulation to incorporate replays and modify the reward function to utilize all data within a listening session. These enhancements improve DQN's hit rate across playlists. While the average hit rate using \textcite{meggetto2023people}'s original reward function is 63\%, our modifications increase it to 78\%, which matches the hit rate obtained by BERT and MC, and falls within 1 percentage point of MLP (79\%) and pMC (77\%).
     \item  The Markov model with position-independent transition probabilities 
     (MC) has only five parameters per playlist, but matches the 78\% average hit rate obtained by BERT and DQN.\footnote{This slightly surpasses the 76\% hit rate for the Transformer when it uses only the predicted remaining time as a feature (Table \ref{tab:overall_accuracy_action}).} The Markov model with position-dependent transition probabilities (pMC) has more parameters than MC yet attains a marginally {\it lower} average hit rate (77\%) than MC. This slight decrease is likely a result of fewer observations for later positions, leading to noisier empirical estimates of the associated transition probabilities. In contrast, the pooling of data across all positions reduces the variance of estimated transition probabilities for MC.
     \item The substantial improvement in both MC and pMC hit rates compared to the zero-order model suggests the presence of state dependence, which is evident even at the aggregate level across sessions and playlists.    
    \begin{table}[!ht]
    \centering
    \caption{Markov chain (MC) transition probabilities  across all playlists.}
    \label{tab:mc_transition_average}
    \begin{threeparttable}
    \begin{tabular}{lccc}
    \hline
    Previous Action &$\mathbb{P}$(skip next song) & $\mathbb{P}$(play next song) & $\mathbb{P}$(replay song)\\
    \hline
    Skip song&  $0.85$  &$0.15$  &\  $0.00^{*}$  \\ 
    Play song& $0.31$  & $0.65$ & $0.03$  \\ 
     Replay song&$0.38$   & $0.62$  &\ \ $0.00^{**}$  \\
    \hline
    \end{tabular}%
    \begin{tablenotes}[para,flushleft]
      \footnotesize
    $^{*}$By definition, a skipped song has a zero probability of being replayed. \\ 
    $^{**}$Value is zero because no song is replayed more than once.  
    \end{tablenotes}
    \end{threeparttable}
    \end{table}
    Table \ref{tab:mc_transition_average}  shows that the probability of playing a song is high (0.65) if the previous song was played, and the probability of skipping a song is still higher (0.85) if the previous song was skipped. The probability of replaying a song once (0.03) is low; and after a replay, the probability of playing the next song (0.62) is high. The transition probabilities in Table \ref{tab:mc_transition_average} also  suggest that MC and pMC achieve comparable holdout hit rates because the highest transition probabilities in each row  are substantially higher than one-half (Table \ref{tab:mc_transition} in Appendix D shows the transition probabilities separately for each playlist). As a result, although MC and pMC have different transition probabilities, they tend to predict the same state-dependent outcomes in the holdout data and obtain similar hit rates.
    \end{enumerate}

\subsection{Hit-rate distributions}
\begin{figure}[!ht]
  \centering
  \includegraphics[width=.8\textwidth]{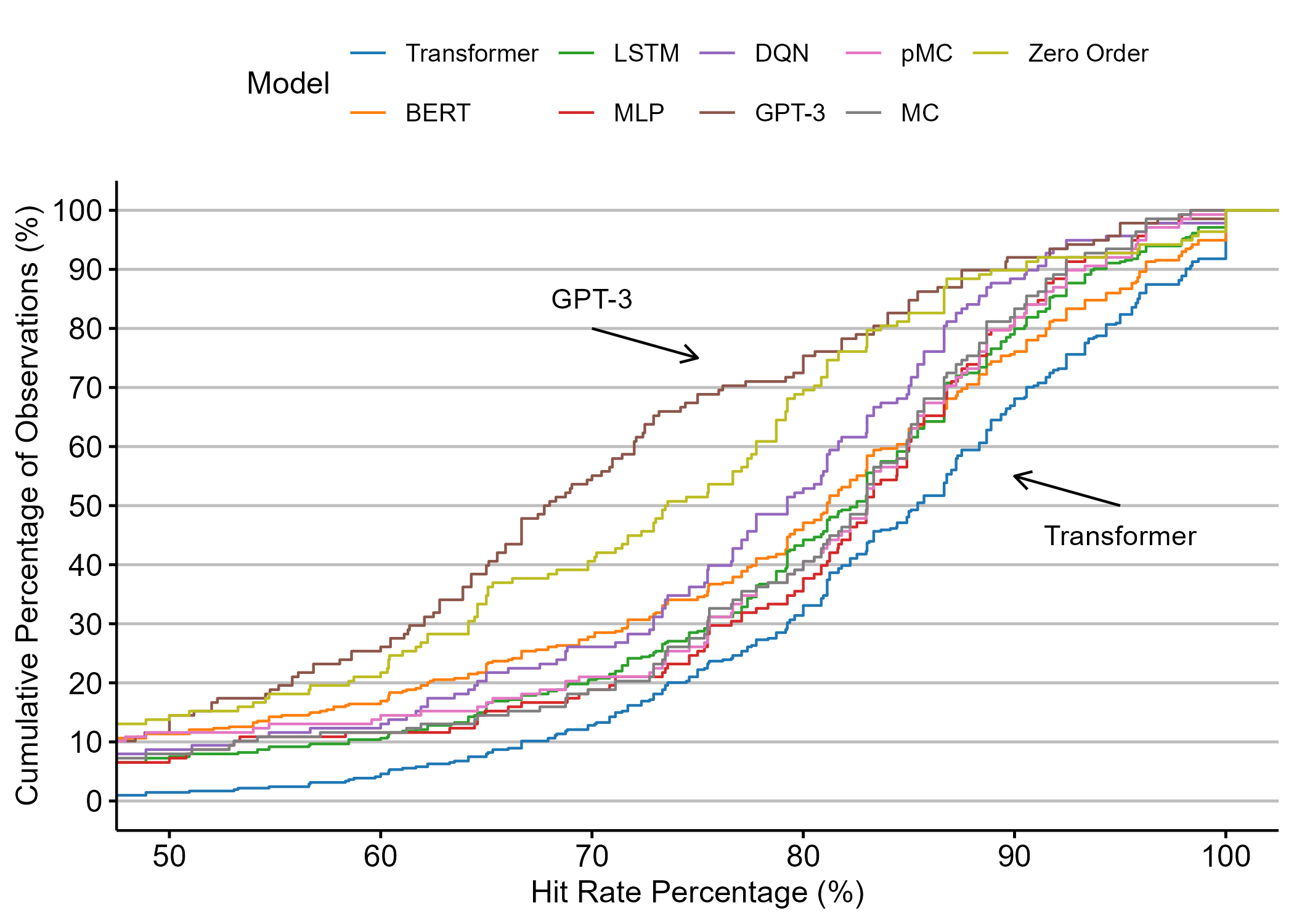}
  \caption{Empirical cumulative distribution functions of hit rates for the nine models.} 
  \label{acc_trans} 
\end{figure}

Figure \ref{acc_trans} shows the empirical cumulative distribution functions (CDFs) of the hit rates for the various models. The distribution functions are based on the average hit rates calculated for each of the 138 positions across the ten playlists.
The Transformer CDF lies below, and the GTP-3 CDF above, all others. 70\% of the Transformer hit rates exceed 80\%, and approximately 30\% surpass 90\%; only 3\% of the Transformer hit rates fall below 60\%. In contrast, 26\% of the GPT-3  hit rates are below 60\%, only 25\% exceed 80\%, and a mere 8\% surpass 90\%. LSTM's performance closely matches the Transformer for hit rates below 70\%, but the gap between their CDFs widens at higher hit rates. The zero-order model consistently outperforms GPT-3, with about 60\% of its hit rates exceeding 70\%, compared to 45\% for GPT-3. MC, pMC, MLP, and BERT exhibit similar hit-rate distributions. DQN shows a marked improvement in the hit-rate distribution compared to both the zero-order model and GPT-3. 

\subsection{Hit rates by type of outcome}
Table \ref{confusion_matrix} shows the confusion matrix --- the hit rates of the different models when the actual outcome is skip, play or replay. All models achieve hit rates above 83\% (and LSTM and the zero-order model obtain the highest hit rate of 89\%) when predicting skips. The zero-order model excels at predicting ``true'' skips because the skip rate is generally high for positions where skipping is the most frequent outcome (see Figure \ref{fig:summary_all}). However, there are substantial differences in the hit rates obtained by the models when predicting plays and replays. The Transformer achieves the highest hit rate of 79\% for plays, followed by MC (73\%) and DQN (72\%). At the other extreme, the zero-order model, GPT-3 and BERT correctly predict plays in only 38\%, 43\%, and 59\% cases, respectively. 
\begin{table}[!ht]
\centering
\caption{Confusion matrix: holdout hit rates for models when the outcome is skip, play or replay.}
\label{confusion_matrix}
\begin{threeparttable}

\begin{tabular}{l|c|c|c|c|c|c|c|c|c}
\hline
\multicolumn{1}{l|}{Outcome $\rightarrow$} & \multicolumn{3}{c|}{Skip} & \multicolumn{3}{c|}{Play} & \multicolumn{3}{c}{Replay} \\
\cline{1-1} \cline{2-4} \cline{5-7} \cline{8-10}
Model prediction $\rightarrow$ & Skip & Play & Replay & Skip & Play & Replay & Skip & Play & Replay\\
\hline
Transformer & 0.86 & 0.13 & 0.01 & 0.21 & 0.79 & 0.00 & 0.12 & 0.00 & 0.88\\
\cline{1-10}
LSTM & 0.89 & 0.11 & 0.00 & 0.33 & 0.67 & 0.00 & 0.45 & 0.00 & 0.55\\
\cline{1-10}
MLP & 0.84 & 0.16 & 0.00 & 0.29 & 0.71 & 0.00 & 0.93 & 0.00 & 0.07\\
\cline{1-10}
DQN  & 0.84 & 0.16 & 0.00 & 0.28 & 0.72 & 0.00 & 0.90 & 0.00 & 0.10\\
\cline{1-10}
BERT & 0.89 & 0.11 & 0.00 & 0.41 & 0.59 & 0.00 & 0.68 & 0.00 & 0.32\\
\cline{1-10}
GPT-3$^{*}$ & 0.84 & 0.09 & 0.06 & 0.57 & 0.43 & 0.00 & 0.82 & 0.00 & 0.18\\
\cline{1-10}
pMC & 0.86 & 0.12 & 0.01 & 0.36 & 0.64 & 0.00 & 1.00 & 0.00 & 0.00\\
\cline{1-10}
MC & 0.83 & 0.17 & 0.00 & 0.27 & 0.73 & 0.00 & 0.99 & 0.00 & 0.01\\
\cline{1-10}
Zero Order$^{**}$ & 0.89 & 0.11 & 0.00 & 0.62 & 0.38 & 0.00 & 0.56 &  0.44 & 0.00\\
\hline
\end{tabular}
\begin{tablenotes}[para,flushleft]
    \footnotesize
$^{*}$The input features exclude remaining time. \\
$^{**}$In a zero-order model, ``play'' means that a song was played at least once, and replay that it was played more than once. 
\end{tablenotes}
\end{threeparttable}
\end{table}

Across all models, on average skips are incorrectly predicted to be plays in 12.8\% cases, and plays incorrectly predicted to be skips in 37.1\% cases. The Transformer achieves an impressive hit rate of 88\%  when predicting replays; the next-best hit rate is 55\% for LSTM, followed by 32\% for BERT, 18\% for GPT-3, and 10\% for DQN. The two Markov models and the zero-order model almost never predict replays correctly. For all models, skip is the dominant incorrect prediction when the true outcome is play or replay. Overall, these results further suggest that the Transformer achieves the best predictions among the tested models.

\subsection{Estimating demand}
We define the demand rate for a song on a playlist as the expected number of times a random playlist listener will play the song. Multiplying the demand rate by the number of playlist listeners gives an estimate of a song's demand from the playlist. 

For each model, we use the predicted listening probabilities for the holdout data to estimate the demand rate for each song on each playlist. (GPT and DQN make deterministic predictions, equivalent to predicting  listening probabilities equal to zero or one.) Note that since the holdout sessions are randomly selected for each playlist, the predicted demand rates for the zero-order model are, by definition, close to the actual demand rates. For this reason, we do not use this model to predict the demand rates. 

Let $y_{i}$ denote the actual demand rate, and ${\hat y_{i}}$ a model's prediction of the demand rate, for song $i$ in the holdout data.  Let ${\bar y}$ be the average demand rate across all songs on all playlists in the holdout data. Let 
$\text{TSS}=\sum_i (y_{i} - \bar{y})^2$ 
denote the total sum of squares and
$\text{SSE}=\sum_i (y_{i} - {\hat y_{i}})^2$
the error sum of squares. Then 
$\text{pseudo}\ R^2 = 1 - ({\text{SSE}}/{\text{TSS}})$ is a measure of the accuracy of the predicted demand rate obtained using a model. Since $y_{i}={\hat y_{i}}$ when there is no prediction error, pseudo $R^2$ can obtain a maximum value of one when a model perfectly  predicts the demand rate for each song. However, pseudo $R^2$ can become negative because (unlike a regression) the error sum of squares for a model is not directly related to the total sum of squares. If a model obtains a negative psuedo $R^2$, we should use the average demand rate rather than the model to predict each song's demand rate. 

\begin{figure}[htbp]
    \centering
        \includegraphics[width=.9\textwidth]{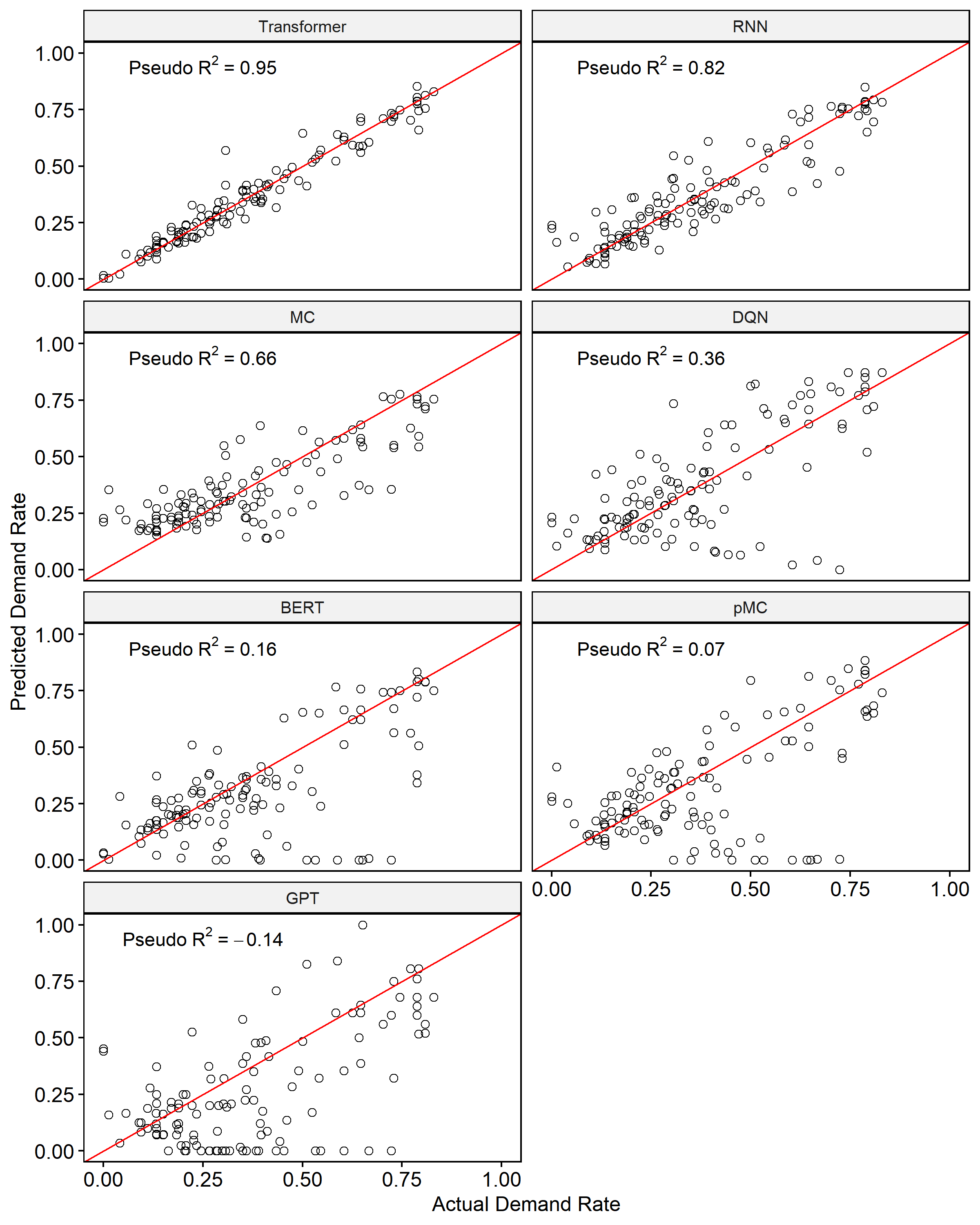}
    \caption{Actual and predicted demand rates for tracks across playlists for each model in the holdout data.}
    \label{fig:demand_p1}
\end{figure}

Figure \ref{fig:demand_p1} shows the actual and predicted demand rates for the 125 tracks across the ten playlists in the holdout data. These predictions exclude the first song for each playlist because, as noted, the models use the outcome for the previous song (play, replay, or skip) as a predictor variable. The estimated demand rate is remarkably accurate for the Transformer (pseudo $R^2=0.95$). The predictions of RNN, MC, DQN, BERT, pMC and GPT become progressively worse, with GPT obtaining a negative value of pseudo $R^2$. 

These results suggest that the Transformer  holds much promise for estimating the demand rate for each item in a bundle. Multiplying an item's demand rate by the number of individuals who purchase a bundle gives the demand for the item from the bundle. If the item appears in multiple bundles, then adding its demand from each bundle gives an estimate of its total demand. In situations where supply is constrained, such as the number of seats in a sporting arena or theater, an accurate estimate of the demand from bundles can be useful for a seller to determine how many individual tickets it should offer for sale for the next game or performance.  For broadcast events, such as baseball games streamed by Amazon Prime, the Transformer can be used to estimate the individual-level probability of viewing the next game, as well as the total viewership for the next game at various levels of aggregation. The streaming service can use these estimates to determine how much effort it should invest in promoting the broadcast, identifying the individuals to whom it should target the promotions, and identify the other complementary content added to boost the viewership of an event.

\subsection{Listening time}
\textcite{meggetto2023people} observed that many predictive models in the Spotify Challenge suffered from a data leak problem because they used the actual listening time of a session --- which cannot be observed until the end of the session --- to predict skipping behavior. They reported sharply lower hit rates when this feature was excluded from their DQN model. Our model avoids the data leak by using the predicted remaining time as a feature instead. To assess how much the predictions would improve if we allowed the data leak, we replaced the predicted remaining time by the observed remaining time for a position (i.e., the total listening time minus the time already listened) as a feature. 
\begin{table}[ht]
\centering
\caption{Average hit rates for models using observed and predicted remaining times as features.}
\label{tab:overall_accuracy_time}
\centering
\begin{tabular}{lcc}
\hline
\multicolumn{1}{l}{\multirow{2}{*}{Model}} & \multicolumn{2}{c}{Average hit rate}                 \\ 
\cline{2-3}
\multicolumn{1}{c}{}                       
& Using predicted & Using observed \\ 
& remaining time & remaining time  \\ 

\hline
Transformer  & 0.84 & 0.91\\
LSTM  & 0.81 & 0.83\\
BERT  & 0.78 & 0.83 \\
MLP   & 0.79 & 0.85 \\
DQN   & 0.78& 0.85 \\
\hline
\end{tabular}
\end{table}

Table \ref{tab:overall_accuracy_time} shows the results. All five models (decoder-only Transformer, LSTM, BERT, MLP and DQN) perform substantially better when the observed remaining time is used as a feature. The hit rate for the Transformer and DQN increases by 7\%, for MLP by 6\%, for BERT by 5\%, and for LSTM by 2\%. One reason for this improvement is that it is trivial to predict  skip when the observed remaining time is less than the length of any remaining song on a playlist. Similarly, it is trivial to predict play or replay for the last position when the observed remaining time is an integer multiple of the length of the last song on a playlist.  
\begin{figure}[!ht]
\centering 
\includegraphics[width=.9\textwidth]{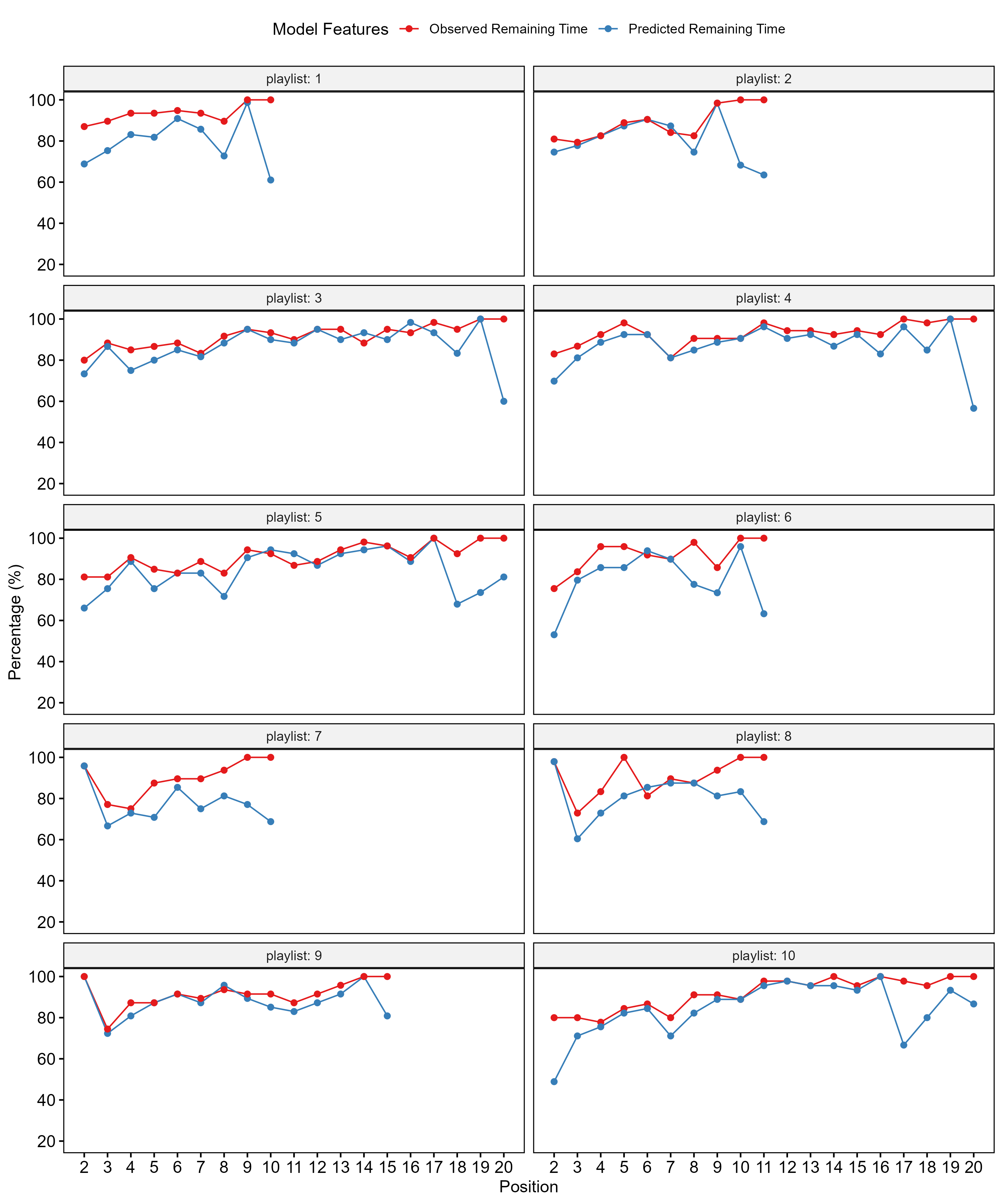}
\caption{Transformer hit rates by position using predicted remaining time and observed remaining time as alternative model features.}
\label{acc_session_length_per_playlist}
\end{figure}
Consistent with this observation, the Transformer achieves a 100\% hit rate for the last position (and for the second-last position) when it uses the observed remaining time as a model feature (see Figure \ref{acc_session_length_per_playlist}). The models obtain comparable hit rates for other positions when predicted or observed remaining time is used as a feature.

\subsection{Transformer attention and listening behavior}
The defining feature of the Transformer is its self-attention mechanism, which is used to predict subsequent outcomes. \textcite{vaswani2017attention} define a \textit{query} as a position for which a prediction is being made, and a \textit{key} as a position that influences the prediction. We analyze the patterns in the self-attention weights of the (decoder) Transformer model. To our knowledge, the following method of analysis is new.






Let $n$ denote the number of positions in a listening session and $\alpha_{ij}$ the self-attention weight for query $i$ and key $j$. As discussed in Appendix E, masked self-attention imposes two constraints:
$$\alpha_{ij} = 0,\ \text{for\ all}\ i+1 \leq j \leq n\ \text{and}\ i = 1, \dots, n,\ \text{and}\ $$
$$\sum_{j \leq i} \alpha_{ij} = 1,\ \text{for\ all}\ i = 1, \dots, n.$$
Let
$${\bar \alpha_{i\cdot}} =\frac{1}{n}\sum_{j=1}^i \alpha_{ij}$$
and
$${\bar \alpha_{\cdot j}} =\frac{1}{n+1-j}\sum_{i=j}^n \alpha_{ij}$$
represent the average non-zero self-attention weights associated with query $i$ and key $j$, respectively.  We call ${\bar \alpha_{i\cdot}}$ the average query weight and ${\bar \alpha_{\cdot j}}$ the average key weight.  The constraint $\sum_{j\le i} \alpha_{ij}=1$ implies that the average query weight is ${\bar \alpha_{i\cdot}}=1/i$, for all $i=1, \dots, n$; that is, later queries have smaller average query weights. In contrast, ${\bar \alpha_{\cdot j}}$ is not subject to a similar constraint and may either increase or decrease with the value of $j$. We compare empirical patterns in ${\bar \alpha_{\cdot j}}$ to the following three baseline models.
\begin{itemize}
    \item[(1)] Baseline 1: Let $\alpha_{ij}=1$ for $j=i$ and $\alpha_{ij}=0$ for all $j\ne i$. That is,  no other position is relevant for predicting the outcome for position $i$. This baseline is similar to a first-order Markov model because the input vector for position $i$ includes the outcome for position $i-1$ as a feature. For this baseline, ${\bar \alpha_{\cdot j}}=1/(n-j+1)$, for all $j=1, \dots, n$, which is an increasing function of $j$. 
    \item[(2)] Baseline 2: Let $\alpha_{ij}=1$ for $j=1$ and $\alpha_{ij}=0$ for all $j\ne 1$. That is, only position $1$ is important for predicting outcome $i$. For this baseline, the average  key weight is ${\bar \alpha_{\cdot j}}=1$ for $j=1$ and ${\bar \alpha_{\cdot j}}=0$, for all $j=2, \dots, n$. 
    \item[(3)] Baseline 3: Let $\alpha_{ij}=1/i$, for all $j\le i$ and $i=1, \dots, n$. That is, all positions $j\le i$ are equally important for predicting outcome $i$. For this baseline, the average  key weight is 
    $${\bar \alpha_{\cdot j}}^0=\frac{1}{n-j+1}\sum_{k=j}^n \frac{1}{k},\ \text{for\ all}\ j=1, \dots, n,$$
    which decreases as $j$ increases from $1$ to $n$.
\end{itemize}
We can construct other baselines for the average key weights, but consider these three because they are the simplest, and as discussed next, baseline 3 is a very good approximation of the empirical average key weights.

\begin{figure}
\centering 
\includegraphics[width=.95\textwidth]{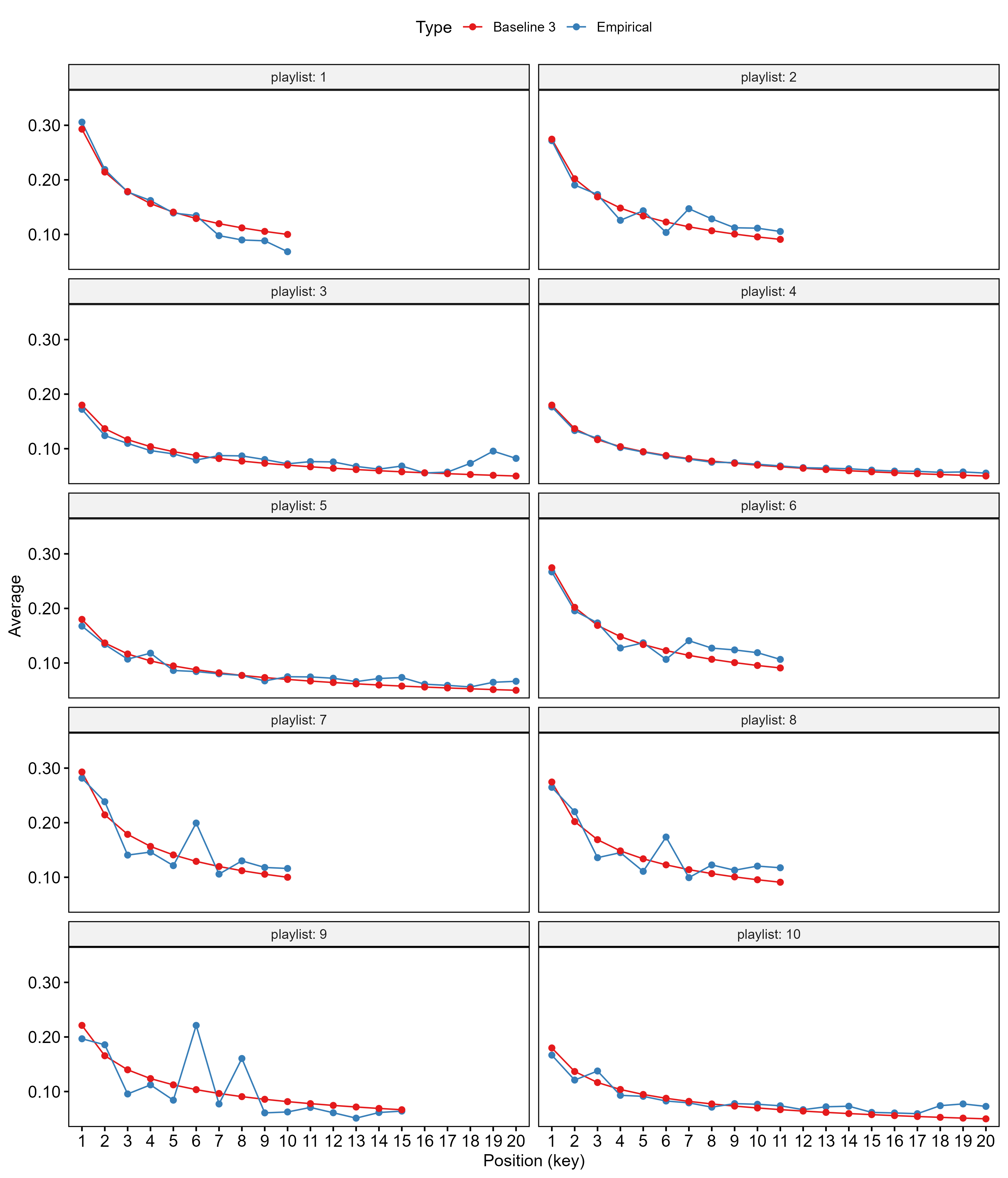}
\caption{Comparisons of empirical (${\bar \alpha_{\cdot j}}$) and baseline (${\bar \alpha_{\cdot j}}^{0}$) values of the average key weights for  playlists.} \label{allplaylist_alpha_bar}
\end{figure}

The self-attention weights can vary over attention heads and attention layers for each listening session. We calculate the average value of ${\bar \alpha_{\cdot j}}$ across the eight attention heads in each layer, and across all three layers, for each session of a playlist. We refer to this value as the empirical average (key weight) ${\bar \alpha_{\cdot j}}$. Given a session, we calculate the values of the empirical average ${\bar \alpha_{\cdot j}}$ and the baseline ${\bar \alpha_{\cdot j}}^0$, for each value of $j$. 
   \begin{table}[!ht]
      \caption{Correlation between the empirical average key weight and baseline 3 for the ten playlists.}
      \label{attention_weights}
      \centering
      \begin{tabular}{lcccccccccc}
        \toprule
        & \multicolumn{10}{c}{Playlist}       \\ \cline{2-11} 
        & 1   & 2   & 3   & 4   & 5   & 6   & 7   & 8   & 9   & 10  \\ \cline{2-11} 
    &0.987&0.988&0.971&0.998&0.981&0.986&0.970&0.974&0.922&0.977\\
      \hline
      \end{tabular}
    \end{table}
Table \ref{attention_weights} reports the correlation coefficient between these two sets of values across sessions for each  playlist. This correlation is consistently high and suggests that baseline 3 is a good approximation to the empirical average key weights. Figure \ref{allplaylist_alpha_bar} displays playlist-specific values of the  empirical average ${\bar \alpha_{\cdot j}}$ (blue) and baseline 3 (red) as a function of $j$. Consistent with the high correlations, the empirical average ${\bar \alpha_{\cdot j}}$ value is close to ${\bar \alpha_{\cdot j}}^0$ for most playlists, especially playlists 3, 4, 5 and 10. But there are also differences for certain position on other playlists, most notably  playlist 9, for which positions 6 and 8 have substantially higher values of the empirical average ${\bar \alpha_{\cdot j}}$ values than the baseline ${\bar \alpha_{\cdot j}}^0$.

\begin{figure}
\centering 
\includegraphics[width=.85\textwidth]{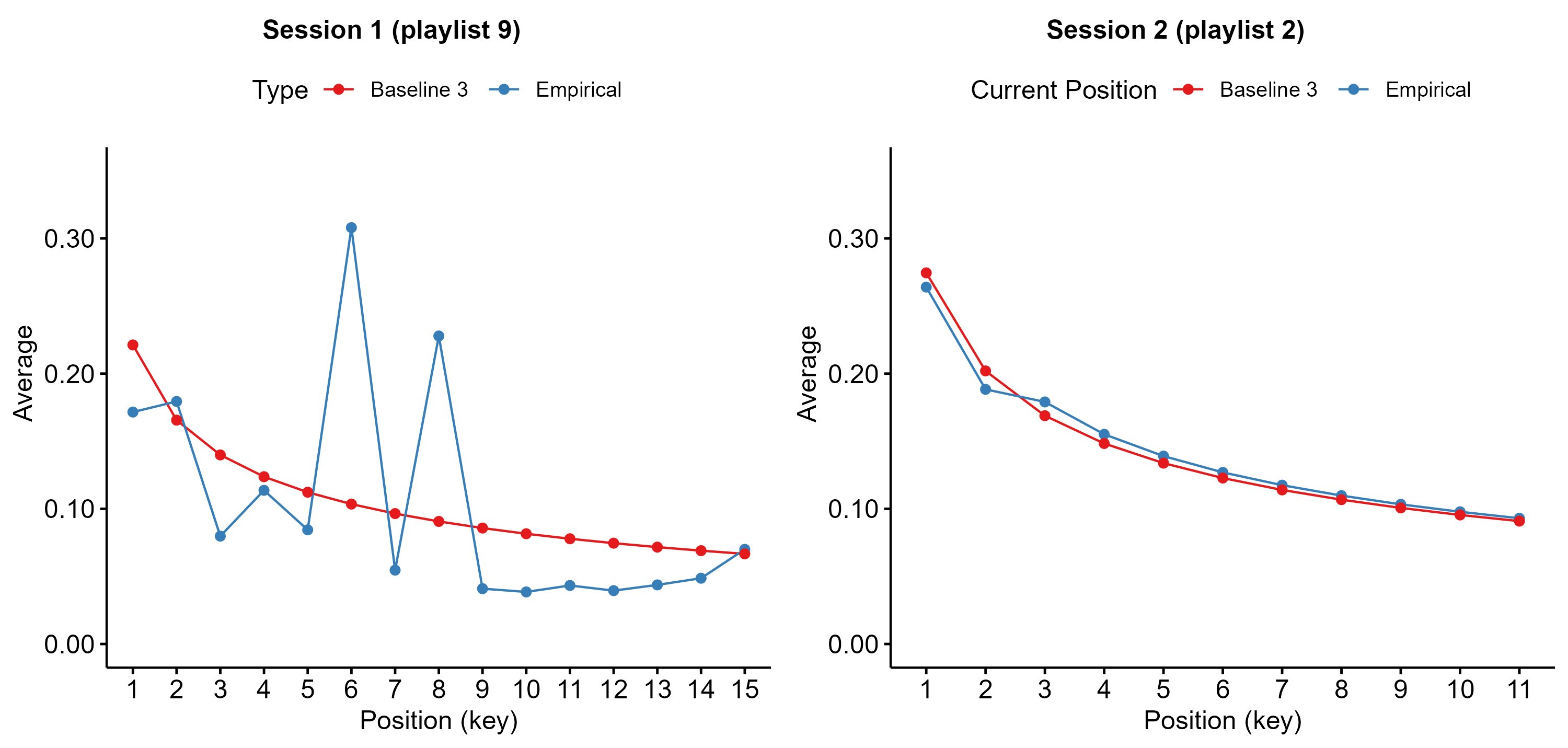}
\caption{Comparisons of average key weights (${\bar \alpha_{\cdot j}}$) and baseline (${\bar \alpha_{\cdot j}}^{0}$) key weights for two  sessions.}
\label{fig:alpha_avg}
\end{figure}

We conclude that the general pattern of key weights is consistent with a Transformer assigning equal weight to each non-succeeding position when predicting an outcome. That said, the strength of the self-attention mechanism is that it allows the attention weights to depend on the specific listening pattern in a session, and we do observe deviations from the general pattern in many sessions. As an illustration, Figure \ref{fig:alpha_avg} shows two sessions, one in which the self-attention weights depart substantially from baseline 3 and the other in which they do not.

\section{Discussion and conclusion}
Experience goods, such as sporting and artistic events, songs, videos, news stories, podcasts, and television series, are often offered and consumed in bundles. A notable feature of these bundles is that the individual items are ordered and can only be consumed sequentially, one at a time. This study examined whether an individual's decision to consume the next item in an ordered bundle can be predicted based on the consumption patterns of the preceding items. 

We evaluated several models to make the prediction, including two custom-built Transformer models that use decoder-only and encoder-decoder architectures, fine-tuned GPT-3, a custom LSTM model, a reinforcement learning model, two Markov models, and a zero-order model. We tested the models using data obtained from Spotify, a music streaming service. We found that the custom-built Transformer with a decoder-only architecture provided the most accurate predictions of individual choices. The hit-rate distribution for this Transformer model stochastically dominated the distributions of all other models.  The same Transformer also provided by far the most accurate predictions for the demand of each item in an ordered bundle.

The individual-level predictions of the custom Transformer could be used  to automatically queue the next song, video, news story, television show, or podcast that an individual is likely to consume from an ordered bundle. For example, a music streaming service like Spotify could use the predictions to determine which track to queue next during an individual listening session. If the model predicts ``play'' or ``replay'', the next song in the queue should remain unchanged. However, if it predicts ``skip'', this prediction should be used as an input to predict the outcome for the following position, the process iterating until ``play'' is predicted, at which point the corresponding song should be  placed next in the queue. Iteratively predicting outcomes is a standard decoder task, which in the present context should halt when ``play'' is generated as an outcome. 

Accurately predicting an item's demand based on its position in a bundle can also be useful for a seller. In situations where supply is constrained, such as the number of seats in a sporting arena or theater, such prediction can assist a seller in determining how many individual tickets to offer for sale for the next game or performance. For broadcast events, such as baseball games, a streaming service can use both individual-level and aggregate demand predictions to assess how much effort it should invest in promoting a particular  broadcast, identify individuals to target with promotions, and identify the other complementary content it might add to boost the viewership of an event.

\textcite{vaswani2017attention} observe that a  Transformer is an auto-regressive model that uses previous outputs as additional input when generating the next output. In the present context, the Transformer captures a general form of state dependence, the next choice depending on the sequence of preceding choices in an ordered bundle. The Transformer attention weights determine the extent to which each preceding choice outcome affects the next one. Our analysis of attention weights suggests that, in the present application, the attention  weights are not concentrated on one or two previous choices but are distributed approximately uniformly across all preceding choices  That said, deviations from this general pattern occur in specific listening sessions.

The Markov model with position-independent transition probabilities (MC) estimated only five parameters to predict whether an individual will replay a song, play the next song, or skip the next song, on a playlist. Its predictions were approximately as accurate as those obtained using pMC, DQN, BERT and MLP, which have hundreds to millions of parameters. MC also predicts aggregate demand for the next item in a bundle much more accurately than any of these models. These results suggest that it is not easy to improve upon the predictions obtained by a simple Markov model, which is also easy to estimate and interpret. The transition probabilities of MC (as well as pMC) suggest that individuals listening to a playlist are either in a listening state or a skipping state: the probability of playing a song is high if the previous song was played, and the probability of skipping a song is higher still if the previous song was skipped. 
As noted, the Transformer model also captures this persistence by using previous outcomes as an input features.

Persistence suggests that a seller offering ordered bundles can benefit by transitioning a user from a non-consuming (skipping) state to a consuming state as early as possible. However, our Transformer model does not provide insights into how to achieve this in the context of musical playlists because (1) each song in a playlist appears in the same fixed position, and (2) none of the song features available in our dataset influence the model's predictions. To go beyond the model, one might consider strategies such as inserting a particularly popular song that is thematically or acoustically similar to the other songs in the playlist to encourage continued listening. Additionally, ensuring that the first item in an ordered bundle is a popular choice can help prevent an individual from entering a skipping state at the outset.

As noted, only previous outcome and predicted remaining time were useful for predicting if a person will skip, play or replay a song on a playlist.  Previous outcome had a substantially greater effect than predicted remaining time on the predictions obtained by the Transformer, LSTM, BERT, MLP, and DQN. The effect of previous outcome is consistent with state dependence. Like other models reported for the Spotify Challenge, song features such as strength, energy, and flatness do not improve predictions in our models. An important exception to this finding is reported by  \textcite{yin2024meta}. Their model uses song features to predict skips and plays for the Spotify data, achieving a hit rate of  73.2\% for the next position on a playlist, which is  lower than the 78\% hit rate obtained by the present Markov model (MC) and substantially lower than the 84\% hit rate obtained by the decoder Transformer. The performance differences could be attributable to three  factors. First, our predictions start from the second position in a session, whereas \textcite{yin2024meta} begin predictions after the sixth position. Second, we focus solely on predicting the next outcome in a listening session, whereas \textcite{yin2024meta} predict a sequence of outcomes. Third, there are differences in the sessions and playlists used in the two studies.

Fine-tuned GPT-3 did not predict well, producing the worst predictions among the nine tested models. There are at least two possible explanations for its poor performance. First, our dataset lacks features such as song title, artist, and listener characteristics, which GPT-3 could potentially leverage from its pre-training to make more accurate predictions. Future research should examine if including such variables can improve GPT’s predictions for other sequential consumption tasks. Second, OpenAI permits the use of only unique listening sequences for fine-tuning GPT-3. This limitation significantly reduces the size of the training sample and prevents the model from learning patterns in repeated sequences. 

Future research could consider several questions related to the design of ordered bundles. In contexts where it is possible to dynamically change the ordering of items in a bundle (such as musical and video playlists, and newsfeed), it can be useful to optimize the ordering of items to maximize (say) the number of items consumed, or the duration over which the items are consumed, by an individual. If the ordering of items cannot be changed (for example, the sequence in which a philharmonic's performances are scheduled), it can be useful to consider the problem of optimizing the (fixed) ordering of items to maximize bundle demand. A related and more difficult problem concerns the optimal selection of both the ordering of a master list of items and the bundles constructed for different segments using this list. Finally, since the ordering of items can affect the value of a bundle for an individual, it can be useful to consider the problem of jointly optimizing the pricing and ordering of items in bundles. 

\theendnotes

\section*{Funding and competing interests}
All authors certify that they have no affiliations with or involvement in any organization or entity with any financial interest or non-financial interest in the subject matter or materials discussed in this manuscript. The authors have no funding to report.

\printbibliography

\newpage

\begin{APPENDICES}

\renewcommand{\thefigure}{\thesection\arabic{figure}}
\renewcommand{\thetable}{\thesection\arabic{table}}
\counterwithin{figure}{section}
\counterwithin{table}{section}


\section{Playlist skip rates by position} 
\label{fig:dist_pl}
Figure \ref{fig:summary_all} shows the skip rate for a track in each position on a playlist. The skip rate is low for the first position on all playlists, after which it increases until reaching a certain position beyond which it generally decreases. 

\begin{figure}[!h]
  \centering
  \includegraphics[width=.9\textwidth]{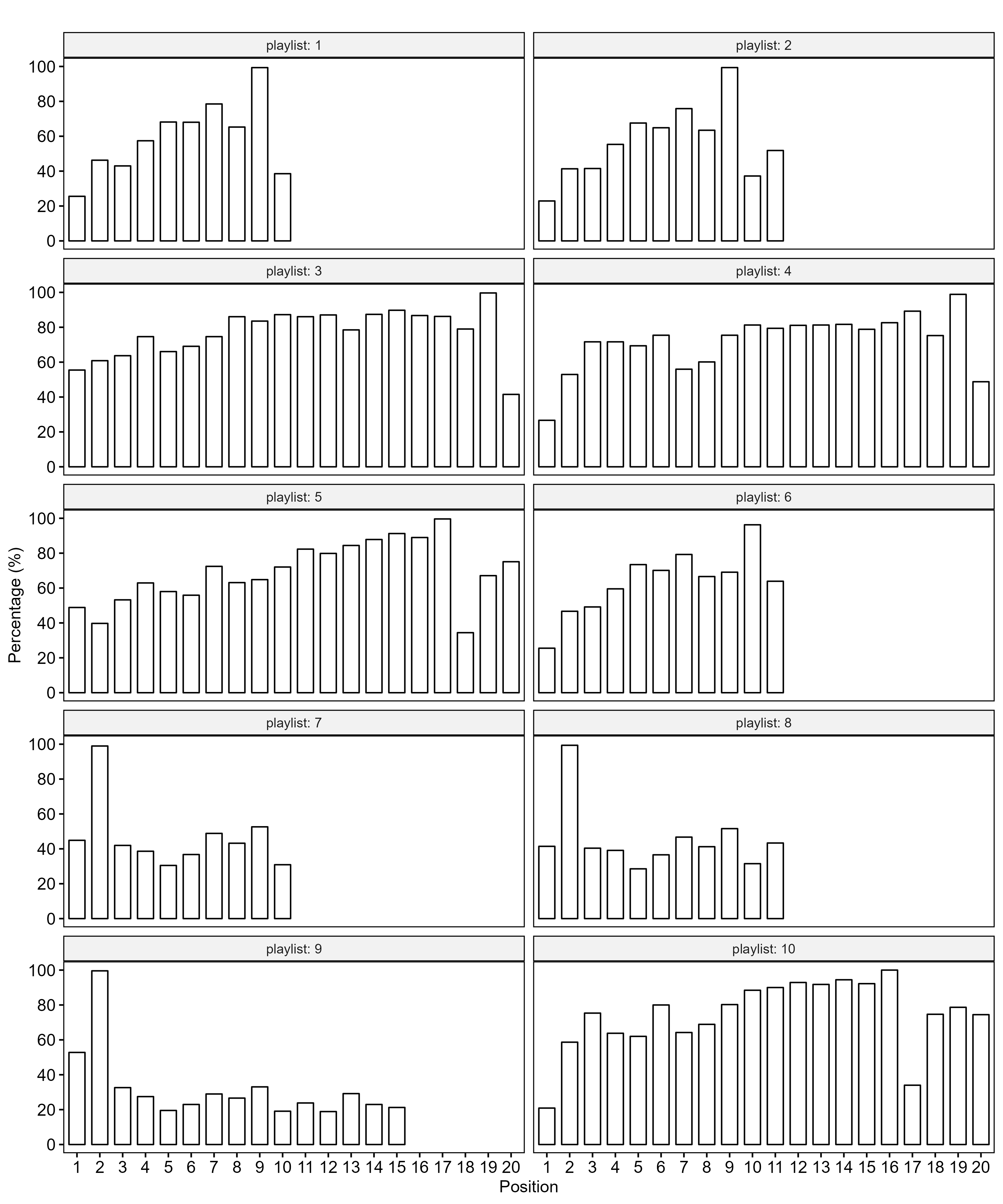}
 \caption{Percentage skip rates by song position for the ten playlists.}
  \label{fig:summary_all}
\end{figure}

\newpage

\section{Summary statistics for training and test playlists.}\label{summary_statistics}

\begin{table}[!htbp] \centering 
\caption{Summary statistics for training playlists.} 
\label{data_description} 
\begin{tabular}{@{\extracolsep{5pt}} rrrrrrrr} 
\hline
Playlist &  
\begin{tabular}[c]{@{}c@{}}\# of \\ tracks \end{tabular}  & 
\begin{tabular}[c]{@{}c@{}}\# of \\ sessions \end{tabular} &
\begin{tabular}[c]{@{}c@{}}Avg. listening\\  time [min:sec]\end{tabular} & 
\begin{tabular}[c]{@{}c@{}}Avg. \# of \\ played songs  \end{tabular} & 
\begin{tabular}[c]{@{}c@{}}Skip [\%]\end{tabular} & 
\begin{tabular}[c]{@{}c@{}} Play [\%]\end{tabular} & 
\begin{tabular}[c]{@{}c@{}} Replay [\%]\end{tabular} \\
\hline
$1$ & $9$ & $686$ & $16:04$ & $3.71$ & $58.80$ & $39.15$ & $2.04$ \\ 
$2$ & $9$ & $566$ & $18:20$ & $3.89$ & $56.83$ & $41.12$ & $2.06$ \\ 
$3$ & $19$ & $535$ & $16:16$ & $4.29$ & $77.40$ & $22.29$ & $0.31$ \\ 
$4$ & $19$ & $476$ & $19:41$ & $5.39$ & $71.62$ & $27.72$ & $0.66$ \\ 
$5$ & $18$ & $473$ & $21:22$ & $5.57$ & $69.08$ & $30.48$ & $0.44$ \\ 
$6$ & $10$ & $433$ & $15:21$ & $3.74$ & $62.65$ & $36.64$ & $0.71$ \\ 
$7$ & $8$ & $431$ & $19:53$ & $4.21$ & $47.38$ & $49.51$ & $3.11$ \\ 
$8$ & $9$ & $425$ & $22:21$ & $4.87$ & $45.88$ & $51.34$ & $2.78$ \\ 
$9$ & $13$ & $419$ & $36:51$ & $8.82$ & $32.17$ & $66.05$ & $1.78$ \\ 
$10$ & $18$ & $405$ & $17:52$ & $4.58$ & $74.58$ & $25.12$ & $0.30$ \\ 
\hline
\hline
Avg.& $13.20$ & $484.90$ & $20:08$ & $4.92$  & $62.68$ &  $36.13$ &  $1.18$ \\ 
\hline
\end{tabular} 
\end{table}

\begin{table}[!htbp] \centering 
\caption{Summary statistics for test playlists.} 
\label{data_description} 
\begin{tabular}{@{\extracolsep{5pt}} rrrrrrrr} 
\hline
Playlist &  
\begin{tabular}[c]{@{}c@{}}\# of \\ tracks \end{tabular}  & 
\begin{tabular}[c]{@{}c@{}}\# of \\ sessions \end{tabular} &
\begin{tabular}[c]{@{}c@{}}Avg.listening\\  time [min:sec]\end{tabular} & 
\begin{tabular}[c]{@{}c@{}}Avg. \# of \\ played songs  \end{tabular} & 
\begin{tabular}[c]{@{}c@{}}Skip [\%]\end{tabular} & 
\begin{tabular}[c]{@{}c@{}} Play [\%]\end{tabular} & 
\begin{tabular}[c]{@{}c@{}} Replay [\%]\end{tabular} \\
\hline
$1$ & $9$ & $77$ & $15:27$ & $3.53$ & $60.78$ & $37.27$ & $1.95$ \\ 
$2$ & $9$ & $63$ & $19:52$ & $4.21$ & $53.25$ & $43.87$ & $2.89$ \\ 
$3$ & $19$ & $60$ & $18:03$ & $4.77$ & $74.92$ & $24.42$ & $0.67$ \\ 
$4$ & $19$ & $53$ & $17:56$ & $4.91$ & $74.15$ & $25.19$ & $0.66$ \\ 
$5$ & $18$ & $53$ & $21:28$ & $5.54$ & $69.25$ & $30.09$ & $0.66$ \\ 
$6$ & $10$ & $49$ & $12:01$ & $2.80$ & $71.99$ & $27.83$ & $0.19$ \\ 
$7$ & $8$ & $48$ & $21:55$ & $4.73$ & $40.83$ & $56.46$ & $2.71$ \\ 
$8$ & $9$ & $48$ & $24:29$ & $5.27$ & $41.48$ & $54.36$ & $4.17$ \\ 
$9$ & $13$ & $47$ & $38:05$ & $9.15$ & $29.65$ & $68.23$ & $2.13$ \\ 
$10$ & $18$ & $45$ & $19:57$ & $5.12$ & $71.56$ & $27.89$ & $0.56$ \\ 
\hline
\hline
Avg.& $13.20$ & $54.30$ & $20:39$ & $5.03$  & $61.90$&  $36.67$ &  $1.42$ \\ 
\hline
\end{tabular} 
\end{table}

\FloatBarrier
\newpage
\section{Hit rates assuming that a session ends after the last played song.}\label{add_acc_trunc}
\begin{table}[ht]
\centering 
\caption{Hit rates assuming that a session ends after the last played song.}
\label{tab:overall_accuracy_truncated}
\resizebox{\textwidth}{!}{%
\begin{tabular}{l|r|c|c|c|c|c|c|c|c|c|c|c}
\hline
\multicolumn{1}{c|}{\multirow{2}{*}{Model}} & \multirow{2}{*}{\begin{tabular}[c]{@{}c@{}} \# of \\ Parameters$^*$ \end{tabular} } & {\multirow{2}{*}{\begin{tabular}[c]{@{}c@{}} Avg. \\ Hit Rate \end{tabular} }} & \multicolumn{10}{c}{Playlist Hit Rate} \\
   \cline{4-13}
  & & & 1 & 2 & 3 & 4 & 5 & 6 & 7 & 8 & 9 & 10\\
\hline
Transformer & 6.8M & 0.84 & 0.82 & 0.82 & 0.86 & 0.86 & 0.84 & 0.76 & 0.80 & 0.82 & 0.89 & 0.82\\
\hline
LSTM & 1.8M & 0.79 & 0.81 & 0.79 & 0.81 & 0.81 & 0.80 & 0.71 & 0.75 & 0.79 & 0.84 & 0.72\\
\hline
BERT & 47.6M & 0.75 & 0.63 & 0.61 & 0.82 & 0.84 & 0.77 & 0.73 & 0.71 & 0.78 & 0.74 & 0.75\\
\hline
MLP & 0.2M & 0.77 & 0.71 & 0.72 & 0.82 & 0.82 & 0.80 & 0.73 & 0.71 & 0.74 & 0.80 & 0.77\\
\hline
DQN & 0.2M & 0.72 & 0.71 & 0.72 & 0.70 & 0.68 & 0.79 & 0.75 & 0.69 & 0.70 & 0.80 & 0.66\\
\hline
GPT-3 & Fine-tuned & 0.74 & 0.61 & 0.78 & 0.49 & 0.73 & 0.88 & 0.64 & 0.79 & 0.75 & 0.75 & 0.88\\
\hline
pMC & [45, 95] & 0.75 & 0.67 & 0.67 & 0.80 & 0.80 & 0.76 & 0.75 & 0.73 & 0.73 & 0.82 & 0.72\\
\hline
MC & 5 & 0.77 & 0.71 & 0.70 & 0.82 & 0.82 & 0.79 & 0.75 & 0.69 & 0.70 & 0.80 & 0.76\\
\hline
Zero Order & [18, 38] & 0.66 & 0.59 & 0.56 & 0.70 & 0.68 & 0.65 & 0.58 & 0.63 & 0.67 & 0.76 & 0.66\\
\hline
\end{tabular} %
}
\end{table}
 
\newpage

\section{Transition probabilities for Markov model (MC) }\label{mc_prob_pl}
\begin{table}[H]
\centering
\caption{Transition probabilities for Markov model (MC) by playlist.}
\label{tab:mc_transition}
\begin{threeparttable} 
 \begin{adjustbox}{center, max totalheight=.95\textheight}
  \begin{tabular}{r|l|c|c|c}
    \hline
    Playlist & Previous Action   & $\mathbb{P}$(skip next song) & $\mathbb{P}$(play next song) & $\mathbb{P}$(replay song)$^{*}$\\
    \hline
      \multirow{3}{*}{1}& Skip & 0.82 & 0.18 & 0.00\\
    \cline{2-5}
     & Play & 0.35 & 0.59 & 0.06\\
    \cline{2-5}
     &Replay$^{**}$& 0.00  & 0.00 & 0.00\\
    \cline{1-5}
     \multirow{3}{*}{2} & Skip & 0.80 & 0.20 & 0.00\\
    \cline{2-5}
     & Play & 0.35 & 0.60 & 0.05\\
    \cline{2-5}
     & Replay  & 0.55 & 0.45 & 0.00\\
    \cline{1-5}
      \multirow{3}{*}{3}& Skip & 0.90 & 0.10 & 0.00\\
    \cline{2-5}
     & Play & 0.37 & 0.62 & 0.02\\
    \cline{2-5}
     & Replay$^{**}$ & 0.00 & 0.00 & 0.00\\
    \cline{1-5}
     \multirow{3}{*}{4} & Skip & 0.88 & 0.12 & 0.00\\
    \cline{2-5}
     & Play & 0.39 & 0.59 & 0.02\\
    \cline{2-5}
     & Replay$^{**}$ & 0.00 & 0.00 & 0.00\\
    \cline{1-5}
     \multirow{3}{*}{5} & Skip & 0.87 & 0.13 & 0.00\\
    \cline{2-5}
     & Play & 0.33 & 0.66 & 0.01\\
    \cline{2-5}
    & Replay & 0.52 & 0.48 & 0.00\\
    \cline{1-5}
     \multirow{3}{*}{6} & Skip & 0.85 & 0.15 & 0.00\\
    \cline{2-5}
     & Play & 0.36 & 0.62 & 0.02\\
    \cline{2-5}
     & Replay$^{**}$ & 0.00 & 0.00 & 0.00\\
    \cline{1-5}
      \multirow{3}{*}{7}& Skip & 0.71 & 0.29 & 0.00\\
    \cline{2-5}
     & Play & 0.28 & 0.65 & 0.07\\
    \cline{2-5}
    & Replay & 0.31 & 0.69 & 0.00\\
    \cline{1-5}
     \multirow{3}{*}{8} & Skip & 0.73 & 0.27 & 0.00\\
    \cline{2-5}
     & Play & 0.26 & 0.68 & 0.06\\
    \cline{2-5}
    & Replay & 0.30 & 0.70 & 0.00\\
    \cline{1-5}
      \multirow{3}{*}{9} & Skip & 0.68 & 0.32 & 0.00\\
    \cline{2-5}
     & Play & 0.13 & 0.84 & 0.03\\
    \cline{2-5}
    & Replay & 0.29 & 0.71 & 0.00\\
    \cline{1-5}
     \multirow{3}{*}{10} & Skip & 0.89 & 0.11 & 0.00\\
    \cline{2-5}
     & Play & 0.44 & 0.55 & 0.01\\
    \cline{2-5}
    & Replay & 0.50 & 0.50 & 0.00    \\
    \hline   
    \multicolumn{5}{l}{ $^{*}$ {\small By definition, a skipped song has a zero probability of being replayed.}} \\
    \multicolumn{5}{l}{ $^{**}$  {\small Probabilities are zero because only the last song was replayed on these playlists. }} 
    \end{tabular}
   \end{adjustbox}




\end{threeparttable}
\end{table}

\FloatBarrier



\newpage
\section{Analysis of Transformer self-attention} \label{appendix_self_attn}
In a Transformer, a self-attention layer maps an input sequence $(\mathbf{x}_1, \dots, \mathbf{x}_n)$ to an output  sequence $(\mathbf{a}_1, \dots, \mathbf{a}_n)$ of the same length. Let $\mathbf{v}_i = \mathbf{x}_i\mathbf{W}^V$, $\mathbf{q}_i=\mathbf{x}_i\mathbf{W}^Q$ and   $\mathbf{k}_i=\mathbf{x}_i \mathbf{W}^K$ denote the value, query and key vectors associated with token $i$, where  $\mathbf{W}^V$, $\mathbf{W}^Q$ and $\mathbf{W}^K$ are weight matrices. 
Let
$$\alpha_{ij} = \text{softmax}\left(\text{score}(\mathbf{x}_i, \mathbf{x}_j)\right),\ \text{for\ all}\ j\le i,$$
denote the self-attention weights, where
$$\text{score}\ (\mathbf{x}_i, \mathbf{x_j})=\frac{\mathbf{q}_i \cdot \mathbf{k}_j}{ \sqrt{d_k}}$$
and $d_k$ denotes the dimension of the key vector $\mathbf{v}_i$. 
The masked self-attention for token $i$ is  
$$\mathbf{a}_i=\sum_{j\le i} \alpha_{ij} \mathbf{v}_j.$$
We can arrange the masked self-attention scores in a matrix 
$$\mathbf{A}=(\mathbf{a}_1, \dots, \mathbf{a}_n)^{'}={\boldsymbol \alpha}\mathbf{v}$$ 
where 
$${\boldsymbol \alpha}=\left(
\begin{array}{cccc}
\alpha_{11} & 0&\cdots &0\\
\alpha_{21} & \alpha_{22}&\cdots &0\\
\vdots & \vdots &\ddots & \vdots \\ 
\alpha_{n1} & \alpha_{n2}&\cdots & \alpha_{nn}
\end{array}
\right)$$
is a lower-triangular matrix of self-attention weights and $\mathbf{v}=(\mathbf{v}_1, \dots, \mathbf{v}_n)^{'}$. 
Observe that $\sum_{j\le i} \alpha_{ij}=1$, for all $i=1, \dots, n$. As a result, the average value of the non-zero self-attention weight  for query $i$ is  
$${\bar \alpha_{i\cdot}}=\frac{1}{i} \sum_{j\le i} \alpha_{ij}=\frac{1}{i}.$$ 
The average self-attention weight  for a key has no constraint but satisfies ${\bar \alpha_{\cdot j}}\le 1$ since $\alpha_{ij}\le 1$, for all $j\le i$ and $i=1, \dots, n$.  The average of the non-zero ${\bar \alpha_{\cdot j}}$ values across the keys is 
$$\frac{1}{n}\sum_j  {\bar \alpha_{\cdot j}}=\frac{1}{n}\left(\frac{1}{ n}\sum_{i\ge 1} \alpha_{i1}+\frac{1}{n-1}\sum_{i\ge 2} \alpha_{i2}+\dots + \frac{1}{1} \alpha_{nn}\right).$$
Consider the baseline $\alpha_{ij}=\alpha_{ij}^0=1/i$, for all $j\le i$ and  $i=1, \dots, n$. 
Then 
$${\boldsymbol \alpha}^0=\left(
\begin{array}{cccc}
1 & 0&\dots &0\\
{1\over 2} & {1\over 2}&\cdots &0\\ 
\vdots & \vdots & \ddots& \vdots \\ 
{1\over n}&{1\over n}&\cdots & {1\over n}
\end{array}
\right)$$
and the average non-zero self-attention weight  for key $j$ is 
$$\alpha_{\cdot j}^0={1\over n-j+1}\left({1\over i}+{1\over i+1}+\dots +{1\over n}\right).$$
The value of $\alpha_{\cdot j}^0$ can be explicit calculated when $n$ is small or $j$ is large (for example, $\alpha_{\cdot j}^0=1/n$ when $j=n$). Otherwise, 
$$\alpha_{\cdot 1}^0={1\over n}\sum_{i=1}^n \alpha_{i1}^0={1\over n}\left(1+{1\over 2}+{1\over 3} + \dots + {1\over n}\right)={1\over n}\left(\ln n +\gamma +{1\over 2n}-\epsilon_n\right)$$
and
$${\bar \alpha_{\cdot j}}^0={1\over n-j+1}\sum_{j\le i} \alpha_{ij}^0={1\over n-j+1}\left(\ln {n \over j-1}+{1\over 2}\left({1\over n}-{1\over j-1}\right)-\left(\epsilon_n - \epsilon_{j-1}\right)\right),\ \text{for}\ j=2, \dots, n,$$
where $\gamma \approx 0.5772$ is the Euler-Masscheroni constant and $0\le \epsilon_n\le 1/8n^2$. The value of ${\bar \alpha_{i\cdot}^0}$ decreases as $i$ increases, and the value of ${\bar \alpha_{\cdot j}}^0$ decreases as $j$ increases. Since the highest value in a sequence of numbers is no smaller than the average value, ${\bar \alpha_{ij}^0}=\alpha_{ij}^0\ge \alpha_{\cdot j}^0$, for all $i=1, \dots, n$. In fact, ${\bar \alpha_{in}^0}=\alpha_{in}^0=\alpha_{\cdot n}^0$  since there is only a single element in column $n$ of ${\boldsymbol \alpha}^0$; otherwise, ${\bar \alpha_{ij}^0}=\alpha_{ij}^0> \alpha_{\cdot j}^0$, for all $i=2, \dots, n$, and ${\bar \alpha_{ij}^0}- {\bar \alpha_{\cdot j}^0}$ decreases monotonically towards zero with increasing query values $i$. However,  ${\bar \alpha_{\cdot j}}$ can increase or decrease with $j$, since there is no constraint on the sum of the column elements in ${\boldsymbol \alpha}$.  
\vfill\eject

\end{APPENDICES}

\end{document}